\documentclass[lettersize,journal]{IEEEtran}
\usepackage{amsmath,amsfonts}
\usepackage{algorithmic}
\usepackage{algorithm}
\usepackage{array}
\usepackage[caption=false,font=normalsize,labelfont=sf,textfont=sf]{subfig}
\usepackage{textcomp}
\usepackage{stfloats}
\usepackage{url}
\usepackage{verbatim}
\usepackage{graphicx}
\usepackage{cite}
\usepackage{multicol}
\usepackage{multirow}
\usepackage{graphicx}
\usepackage{lipsum}
\usepackage{booktabs}
\usepackage{xcolor}
\usepackage{pdfpages}
\title{Using KL-Divergence to Focus Frequency Information in Low-Light Image Enhancement}

\author{%
	Yan~Xingyang, Huang~Xiaohong\textsuperscript{*}, Zhang~Zhao, You~Tian and Xu~Ziheng%
	\thanks{Yan~Xingyang, Huang~Xiaohong, Zhang~zhao, You~Tian and Xu~Ziheng, Huang Xiaohong is the corresponding author, are with the Hebei Key Laboratory of Industrial Intelligent Perception and the College of Artificial Intelligence, North China University of Science and Technology, Tangshan, Hebei 063210, China (e-mail: 
	yxyhh@stu.ncst.edu.cn; Huangxh@ncst.edu.cn; zhaozhang@ncst.edu.cn; tianyou@stu.ncst.edu.cn; xuziheng@stu.ncst.edu.cn)}%
}

\begin{document}

\maketitle


\begin{abstract}
In the Fourier domain, luminance information is primarily encoded in the amplitude spectrum, while spatial structures are captured in the phase components. The traditional Fourier Frequency information fitting employs pixel-wise loss functions, which tend to focus excessively on local information and may lead to global information loss. In this paper, we present LLFDisc, a U-shaped deep enhancement network that integrates cross-attention and gating mechanisms tailored for frequency-aware enhancement. We propose a novel distribution-aware loss that directly fits the Fourier-domain information and minimizes their divergence using a closed-form KL-Divergence objective. This enables the model to align Fourier-domain information more robustly than with conventional MSE-based losses. Furthermore, we enhance the perceptual loss based on VGG by embedding KL-Divergence on extracted deep features, enabling better structural fidelity. Extensive experiments across multiple benchmarks demonstrate that LLFDisc achieves state-of-the-art performance in both qualitative and quantitative evaluations. Our code will be released at: https://github.com/YanXY000/LLFDisc

\end{abstract}


\section{Introduction}
\IEEEPARstart{L}{ow-light} image enhancement (LLIE) is a fundamental task in computer vision, with applications in surveillance, autonomous driving, and computational photography. Recent deep learning approaches typically either extract spatial-domain features or adopt Retinex-based decomposition into reflectance and illumination. However, spatial-domain feature extraction is prone to noise amplification, while Retinex-based methods may suffer from texture distortion and poor generalization across varying illumination conditions.

\begin{figure}[!t]
	\centering
	\includegraphics[width=3.5in,height=1.9in]{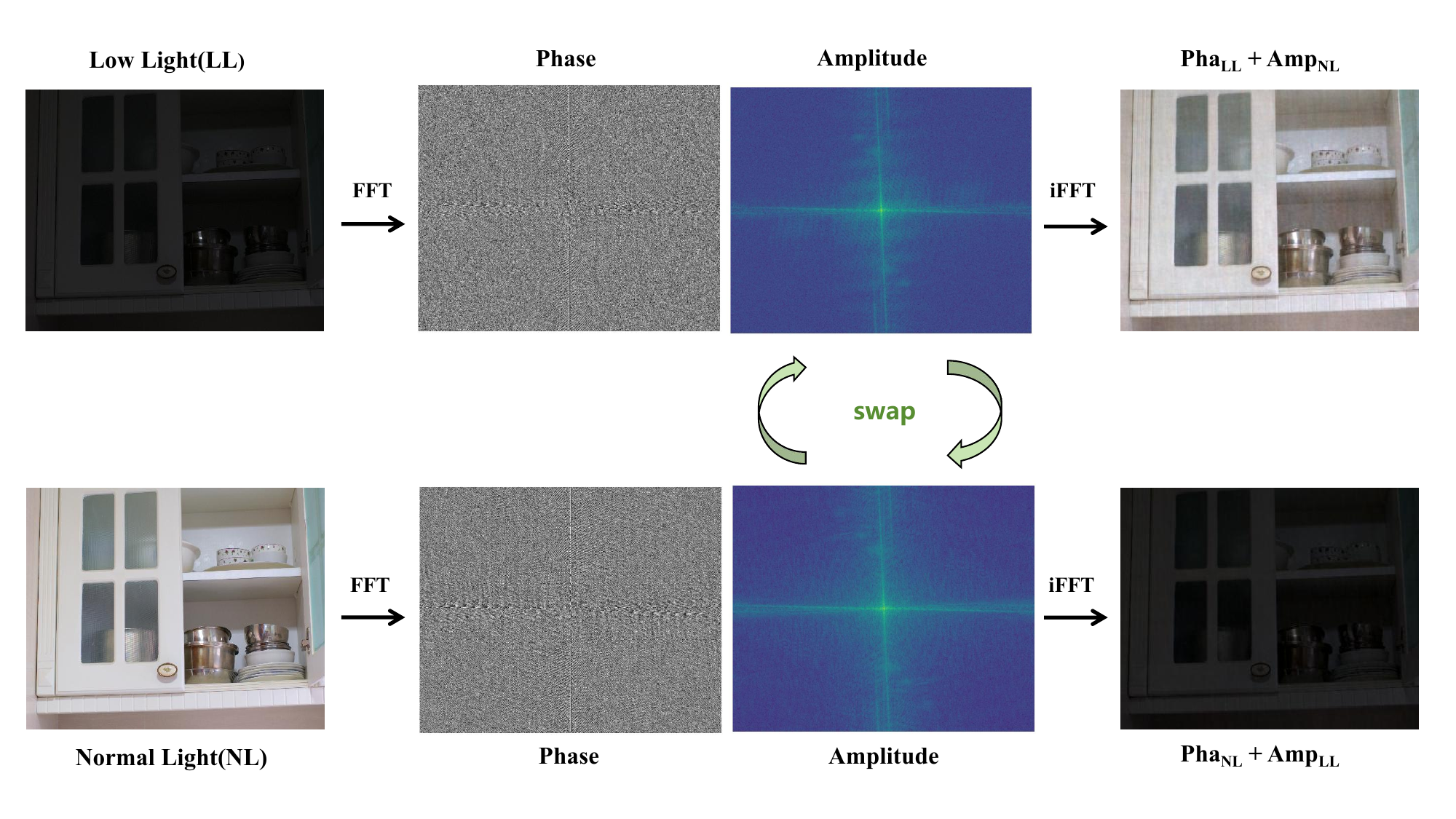}
	\caption{Low-light and normal-light images are Fast Fourier Transformed (FFT) to obtain amplitude and phase information in the frequency domain. By keeping the phase information unchanged and exchanging the amplitude information between the two images, an inverse Fast Fourier Transform (iFFT) is performed to reconstruct the spatial-domain image, which is then visualized to observe the effect.}
	\label{fig:fig2}
\end{figure}

To address these limitations, we explore the Fourier domain, which provides a complementary representation of image formation beyond spatial analysis. Motivated by recent Fourier-based LLIE studies \cite{fourier_LLapp2,fourier_LLapp3,fourier_LLapp4}, we examine the relationship between amplitude spectra and perceived brightness. As illustrated in Fig.~\ref{fig:fig2}, swapping the amplitude spectra between low-light and normal-light images—while preserving phase—produces a reconstruction that closely resembles normal illumination, underscoring the critical role of amplitude in LLIE. Nevertheless, despite its importance, current Fourier-based approaches may not fully exploit amplitude information due to limitations in their loss design.

This limitation often stems from the choice of optimization objectives. Existing Fourier-based enhancement methods typically minimize pixel-wise errors such as Mean Squared Error (MSE) between predicted and target image \cite{fourier_app3,fourier_LLapp2,fourier_app5}. While simple, such losses treat each frequency bin independently and neglect the joint distributional structure across frequencies. In contrast, distribution-based losses model continuous spectral distributions, such as KL-Divergence, potentially capturing richer cross-frequency relationships than pixel-wise objectives. Such objectives have demonstrated strong potential in other vision tasks, suggesting their applicability to LLIE. Motivated by this intuition, we investigate whether KL-Divergence can better fit Fourier-domain information than MSE. As shown in our experiments, KL-based losses consistently outperform MSE-based losses in both quantitative metrics and perceptual quality.

\begin{figure*}[h]
	\includegraphics[height=0.4\textheight,width=\linewidth]{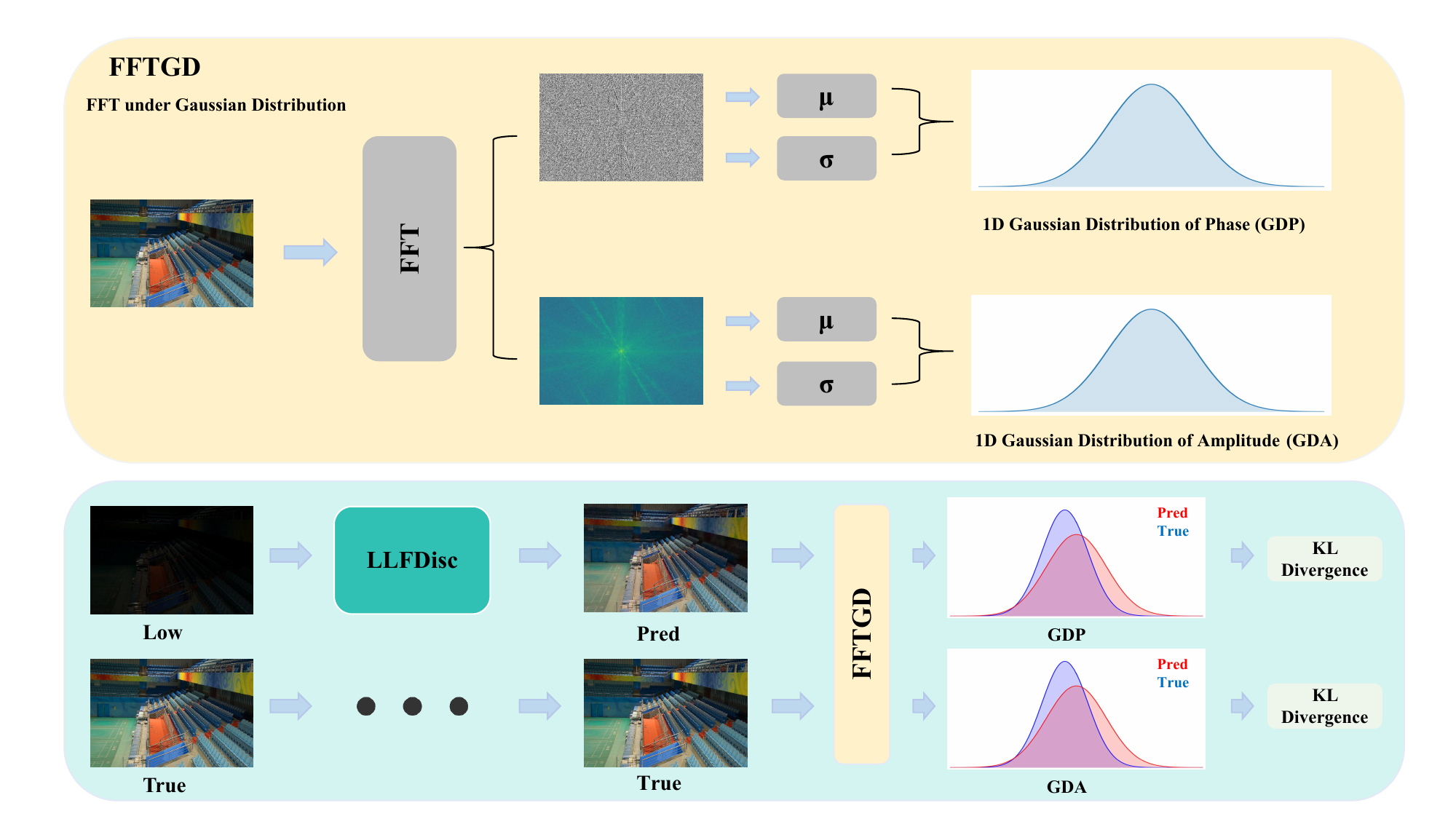}
	\caption{By applying the Fast Fourier Transform (FFT) to images, the amplitude and phase are obtained, and their means ($\mu$) and variances($\sigma$) are calculated. These statistics are used to project the amplitude and phase into Gaussian distributions. Finally, the KL-Divergence between the Gaussian distributions of the predicted and ground-truth images' amplitude and phase is computed and returned as the loss value.}
	\label{fig:fig1}
\end{figure*}

Inspired by distributional modeling in object detection \cite{idea1,idea2}, where bounding box coordinates are projected into Gaussian distributions for KL-based supervision, we propose a Fourier KL loss. Specifically, we model the amplitude and phase spectra of both predicted and ground-truth images as Gaussian distributions parameterized by their means and variances, and compute the KL-Divergence between them as the training objective (Fig. \ref{fig:fig1}). This design explicitly aligns the holistic frequency-domain distribution rather than treating frequency bins independently.

Given the advantage of KL-Divergence in capturing distribution-level relationships, we further extend its application to the spatial perceptual domain. While the Fourier KL 

\noindent loss effectively models frequency-domain distributions, it does not explicitly enforce perceptual consistency in the spatial domain. Therefore, we integrate KL-Divergence into the VGG perceptual loss, enabling it to measure distributional similarity between high-level feature representations. We also present an efficient single-branch network architecture tailored for these losses.

Extensive experiments demonstrate that our method consistently outperforms existing MSE-based Fourier losses, achieving state-of-the-art results on public LLIE benchmarks. To the best of our knowledge, this is the first work to apply KL-Divergence to directly fit Fourier-domain information.

\textbf{Our main contributions are:}
\begin{itemize}
	\item We propose a Fourier-domain loss based on KL-Divergence, achieving state-of-the-art (SOTA) performance on low-light datasets and outperforming previous Fourier-based LLIE methods.
	\item We enhance the existing VGG perceptual loss by incorporating KL-Divergence, thereby significantly improving the representation capability of the model.
	\item We present a streamlined network design that is simple and elegant compared to many existing methods. It consists of only a single branch, thereby reducing the number of parameters and improving efficiency.
\end{itemize}

\section{Related Work}
\textbf{Low-Light Image Enhancement.} There are numerous methods for low-light image enhancement based on deep learning. For example, RetinexNet \cite{LLIE_1} and RetinexFormer \cite{LLIE_2} employ Retinex theory to decompose images into reflectance and illumination components, thereby achieving image enhancement. Uretinex-net \cite{ret_3} employs a U-shaped network combined with Retinex to achieve low-light image enhancement. Diff-Retinex \cite{LLIE_3} by combining Retinex with diffusion models,further addresses content loss and color bias caused by low-light conditions using a generative approach. LYT \cite{LLIE_4} and Bread \cite{LLIE_5} project the color space into the YCbCr space, decouple luminance and color information, and enhance brightness while suppressing noise. CID-Net \cite{LLIE_6} proposes a novel color space, HVI, to tackle low-light image enhancement by decoding color and brightness information, combining the polarized HS axis with trainable intensity to improve image brightness and eliminate color space noise caused by the HSV space. However, these methods require additional transformations and decompositions of the image, which can complicate implementation. In contrast, the RGB color space does not require extra operations and is simpler to implement. Moreover, RGB-based enhancement often yields better results. LLFormer \cite{LLIE_7} and Enlightengan \cite{LLIE_8} have demonstrated effective low-light image enhancement within the RGB space, achieving satisfactory outcomes.

\textbf{Fourier Frequency Information.} Fourier Frequency information has achieved significant success in many other areas of computer vision \cite{fourier_app1}, \cite{fourier_app2}, \cite{fourier_app3}, \cite{fourier_app4, fourier_app5, fourier_app6}.Meanwhile, Fourier frequency information has also demonstrated remarkable effectiveness in low-light image enhancement tasks \cite{fourier_LLapp1}, \cite{fourier_LLapp2}, \cite{fourier_LLapp3}, \cite{fourier_LLapp4}. For example, \cite{fourier_LLapp4} utilized cross-modal infrared images to guide the structural information of the phase components and employed luminance maps to precisely enhance 

\begin{figure*}[h]
	\includegraphics[width=\linewidth]{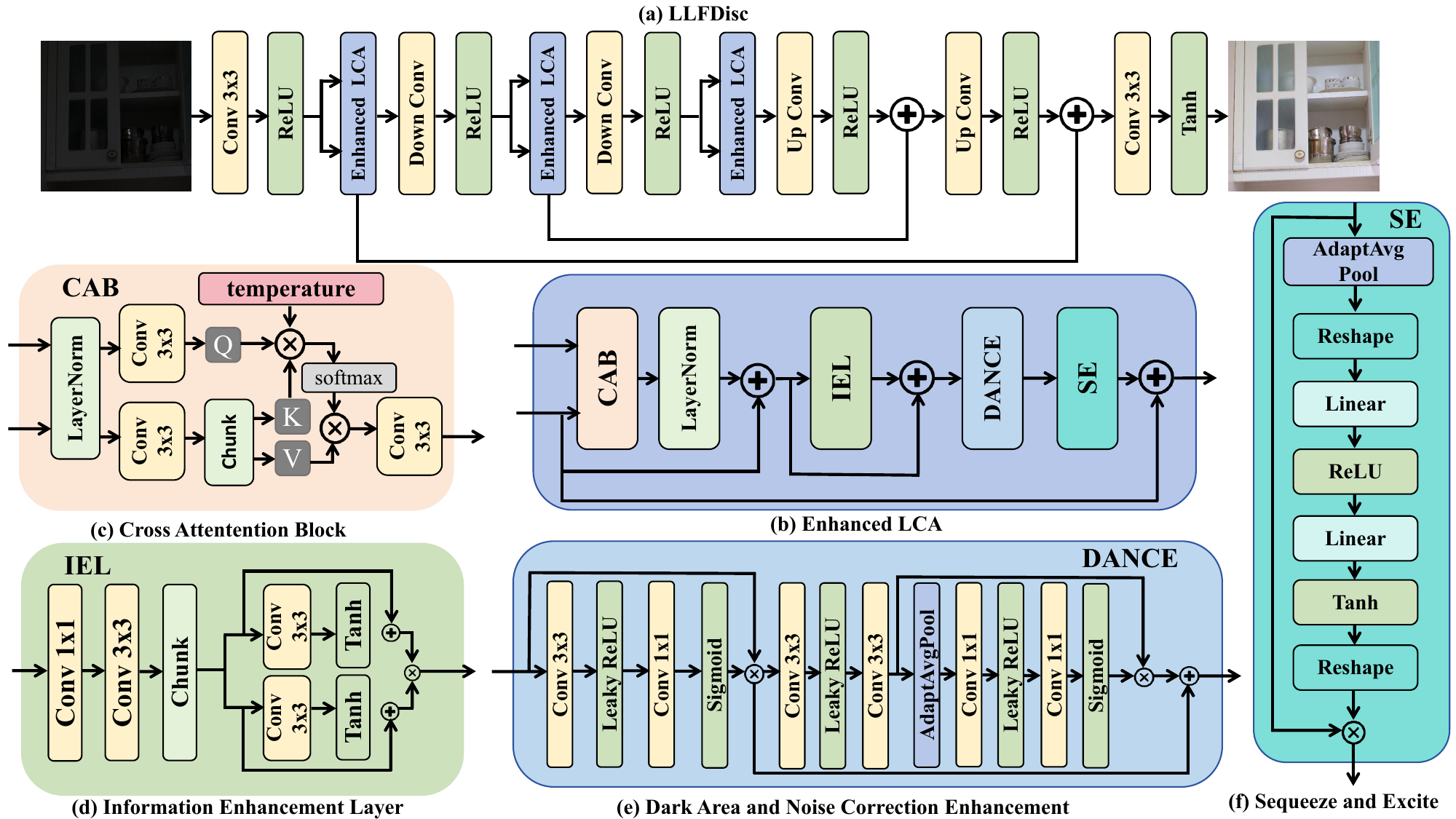}
	\caption{The LLFDisc network is illustrated in the figure, with the upper part depicting the overall structure and the lower part detailing the individual modules. The LLFDisc network enhances low-light images by employing a series of convolutional, activation, downsampling, and upsampling operations, complemented by the Enhanced Lighten Cross Attention (EnhancedLCA) module. These components work synergistically to improve the brightness and detail of low-light images.}
	\label{fig:fig3}
\end{figure*}

\noindent the amplitude components, thereby achieving low-light image enhancement. Xu et al. \cite{fourier_app2} extracted more robust semantic information from the Fourier frequency domain for domain generalization in data augmentation. Lv et al. \cite{fourier_LLapp1} used the amplitude obtained from the Fourier transform as a brightness prior to guide model learning. Hemkant Nehete et al. \cite{fourier_app3} swapped image amplitudes to achieve rain and haze removal. These studies highlight the broad applicability of Fourier frequency information and further motivate its exploration in low-light image enhancement.To ensure model accuracy, inspired by the application of Fourier transforms in low-light image enhancement, Wang et al. \cite{fourier_LLapp2} transformed images into the frequency domain and computed losses based on the amplitude and phase of the resulting spectra for image enhancement. However, these methods only consider pixel wise relationships and are limited in their ability to represent Fourier frequency information comprehensively.

\textbf{KL-Divergence.} KL-Divergence, a fundamental measure of discrepancy between probability distributions in terms of entropy, has proven effective in a wide range of domains, including diffusion models \cite{IG_1} and Variational Autoencoders (VAEs) \cite{IG_2}. Beyond generative modeling, recent works have leveraged KL-Divergence in vision tasks such as object detection \cite{idea1,idea2}, where bounding boxes are modeled as Gaussian distributions and their pairwise distances are computed in distribution space. Building on the distribution-based representation strategy in \cite{idea1}, we propose to incorporate KL-Divergence into the Fourier domain, guiding the model to align the frequency distributions of predictions and ground truth. This design explicitly encourages the network to capture frequency-aware features, enabling more faithful reconstruction in low-light image enhancement.

\section{Proposed Method}
The LLFDisc (\textbf{L}ow \textbf{L}ight \textbf{F}ourier with \textbf{D}ANCE, \textbf{I}EL, \textbf{S}E, \textbf{C}AB) is a deep learning architecture specifically developed for low-light image enhancement. As illustrated in Fig. \ref{fig:fig3}, it adopts a U-shaped encoder–decoder structure composed of three feature-extraction stages, each integrating an Enhanced Lighten Cross-Attention (EnhancedLCA) module. The processing pipeline follows a downsampling–processing–upsampling paradigm. First, the input image is processed by an initial convolution to extract basic features, followed by two downsampling operations to produce multi-scale feature representations. At each scale, the EnhancedLCA module refines features through: (i) a Cross-Attention Block, (ii) an Information Enhancement Layer (IEL) \cite{LLIE_6}, (iii) a Detail-Aware and Noise Correction Enhancement (DANCE) module, and (iv) a Squeeze-and-Excitation (SE) module \cite{se}. Finally, multi-scale features are upsampled using transposed convolution and fused to generate the enhanced output image. The subsequent sections describe the EnhancedLCA module and loss-function design in detail.

\subsection{Preliminary}
Initially, a concise introduction to Fourier frequency-domain information is necessary. Given an input image $X$ with dimensions $H \times W$, a specific transformation function $F$ can be employed to convert the image $X$ from the spatial domain to the Fourier domain. This transformation function $F$ can be mathematically expressed as follows:

{\small
\begin{equation}
	\label{Fourier1}
	\mathcal{F}(x)(u,v) = X(u,v) = \frac{1}{\sqrt{HW}} \sum_{h=0}^{H-1} \sum_{w=0}^{W-1} x(h,w) e^{-j2\pi \left( \frac{h}{H}u + \frac{w}{W}v \right)}
\end{equation}
}

The variables \( u \) and \( v \) represent the frequency coordinates in the Fourier domain, while \( j \) denotes the imaginary unit. This formula describes the process of transforming an image \( x \) from the spatial domain into a complex representation \( F(u, v) \) in the Fourier domain. Here, \( h \) and \( w \) signify the discrete points in the spatial domain, and \( X(u, v) \), which contains complex values, can be expressed using the following formula:

\begin{equation}
	\label{Fourier2}
	\small 
	X(u, v) = \text{R}(X(u, v)) + j \text{I}(X(u, v))
\end{equation}

In the Fourier space, the image \( X(u, v) \) can be decomposed into its amplitude \( A(X(u, v)) \) and phase \( P(X(u, v)) \). The decomposition formulas are as follows:

\begin{equation}
	\label{Fourier3}
	\small 
	\mathcal{A}(X(u,v)) = \sqrt{R^2(X(u,v)) + I^2(X(u,v))}
\end{equation}

\begin{equation}
	\label{Fourier4}
	\small 
	\mathcal{P}(X(u,v)) = \arctan\left[\frac{I(X(u,v))}{R(X(u,v))}\right]
\end{equation}

\subsection{EnhancedLCA}
EnhancedLCA is an improvement upon LCA and comprises several components, including the DANCE, IEL, SE, and CAB modules. Among these, the DANCE module is a dark area and noise correction enhancement module proposed by us. The following sections provide detailed descriptions of each module.

\textbf{DANCE.} The DANCE module integrates three sub-modules to enhance image quality: the Noise Perception Module, the Dark Area Enhancement Module, and the Channel Attention Module. Each module is designed to address specific aspects of image enhancement.
	
	\subsubsection{Noise Perception Module}
	This module suppresses noise using a lightweight architecture. The process begins with a depthwise separable convolution to maintain spatial context while reducing computational load:
	\begin{equation}
		\mathbf{Z}_0 = \text{DWConv}_{3\times3}(\mathbf{X}),
	\end{equation}
	where $\mathbf{X} \in \mathbb{R}^{B \times C \times H \times W}$ is the input feature tensor. This is followed by a LeakyReLU activation function and a $1 \times 1$ pointwise convolution to fuse channel information:
	\begin{equation}
		\mathbf{Z}_1 = \text{Conv}_{1\times1}\left(\text{LeakyReLU}(\mathbf{Z}_0)\right),
	\end{equation}
	culminating in a Sigmoid function that generates a noise suppression weight map:
	\begin{equation}
		\mathbf{M}_{\text{noise}} = \sigma(\mathbf{Z}_1),
	\end{equation}
	where $\sigma$ denotes the sigmoid function. The noise-suppressed feature is obtained by channel-wise modulation:
	\begin{equation}
		\mathbf{X}_{\text{den}} = \mathbf{X} \odot \mathbf{M}_{\text{noise}}.
	\end{equation}
	
	\subsubsection{Dark Area Enhancement Module}
	This module focuses on restoring details in low-light regions using a two-layer convolution structure. The first layer extracts features with a $3 \times 3$ kernel and a LeakyReLU activation:
	\begin{equation}
		\mathbf{F}_0 = \text{LeakyReLU}(\text{Conv}_{3\times3}(\mathbf{X}_{\text{den}})),
	\end{equation}
	while the second layer refines these features without an activation function to preserve original information:
	\begin{equation}
		\mathbf{F}_{\text{dark}} =  \text{Conv}_{3\times3}(\mathbf{F}_0).
	\end{equation}
	This enhances visibility in dark areas without amplifying noise.
	
	\subsubsection{Channel Attention Module}
	This module compresses spatial information into channel descriptors via global average pooling:
	\begin{equation}
		\mathbf{s} = \text{GAP}(\mathbf{F}_{\text{dark}}),
	\end{equation}
	where $\text{GAP}$ denotes global average pooling. It then reduces and expands dimensions through convolutions, followed by a LeakyReLU activation and a Sigmoid function to generate channel-wise weights:
	\begin{equation}
		\mathbf{w} = \sigma\left(\text{Conv}_{1\times1}\left(\text{LeakyReLU}(\text{Conv}_{1\times1}(\mathbf{s}))\right)\right).
	\end{equation}
	This re-weights features, highlighting important channels.
	
	\subsubsection{Forward Summary}
	During forward propagation, input features are processed sequentially through these sub-modules. The Noise Perception Module reduces noise, the Dark Area Enhancement Module enhances low-light details, and the Channel Attention Module re-weights features. The final output combines enhanced and noise-reduced features, resulting in improved image quality with better detail and reduced noise:
	\begin{equation}
		\mathbf{Y} = \mathbf{X}_{\text{den}} + \mathbf{F}_{\text{dark}} \odot \mathbf{w}.
	\end{equation}

\textbf{IEL.}
The IEL module is a convolutional neural network structure that enhances feature selectivity through a gating mechanism. The input first undergoes feature extraction via convolutional layers, followed by processing in a block unit. The gating mechanism, comprising parallel convolutional layers and a Tanh activation function, produces a gating signal. This signal is element-wise multiplied with the block unit's output to dynamically modulate feature flow. This adaptive enhancement or suppression of specific features improves the model's ability to capture important information and inhibits irrelevant or noisy features, thereby boosting the network's expressive power and generalization capability.

\textbf{SE.}
This module is the "Squeeze and Excite" mechanism. It first aggregates global information of the input features using Adaptive Average Pooling (AdaptAvgPool). Subsequently, the features undergo two linear transformations and activation functions (ReLU and Tanh). Finally, the shape is adjusted through reshaping to re-weight the feature channels, thereby enhancing the model's representational capability.

\textbf{CAB.} In this study, the Cross-Attention Block (CAB) enhances dark area details by selectively fusing features using a Query and Key-Value Pair mechanism. It handles large-scale illumination inhomogeneity through a multi-head attention mechanism, with a learnable temperature parameter dynamically adjusting attention weights for various low-light scenarios. Different convolution operations reduce model parameters and expand the receptive field to capture broader spatial information.

\subsection{Fourier KL-Divergence Loss}
Traditional Fourier frequency-domain information fitting employs Mean Squared Error (MSE) for approximation \cite{fourier_app3, fourier_LLapp2, fourier_app5}. However, the use of loss functions such as Mean Squared Error (MSE) to directly measure pixel-wise differences in the Fourier spectral domain can lead to a model that focuses excessively on local information. This approach often neglects the relationships between neighboring pixels during the Fourier domain information reconstruction process. To address this issue, we propose a KL (Kullback-Leibler) divergence-based loss function that operates on the distribution 

\begin{figure*}[h]
	\centering
	\setlength{\tabcolsep}{2pt}
	\begin{tabular}{cccccccc}
		\includegraphics[width=0.115\textwidth, height=0.15\textheight]{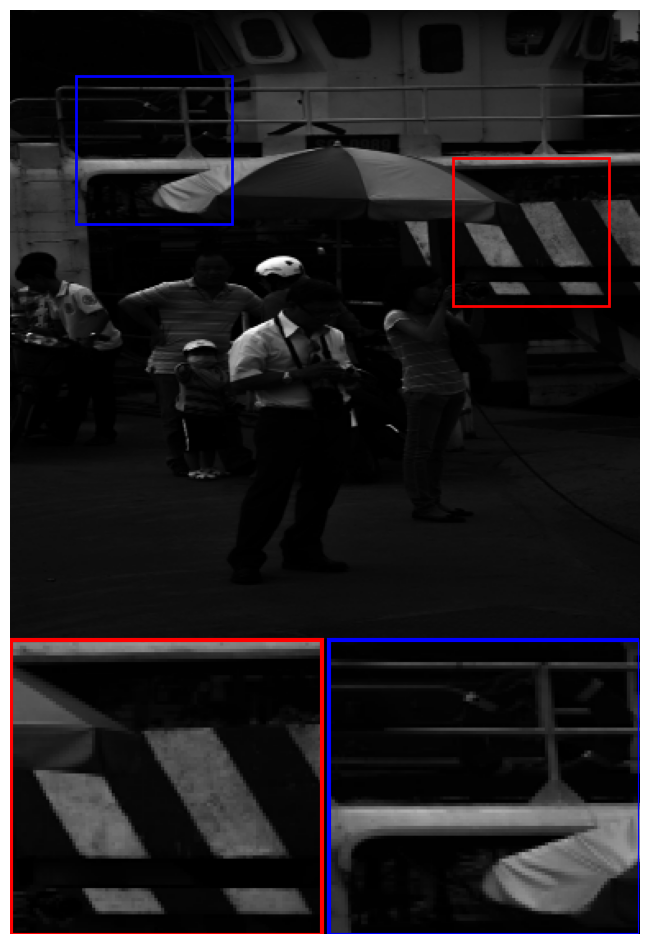} & 
		\includegraphics[width=0.115\textwidth, height=0.15\textheight]{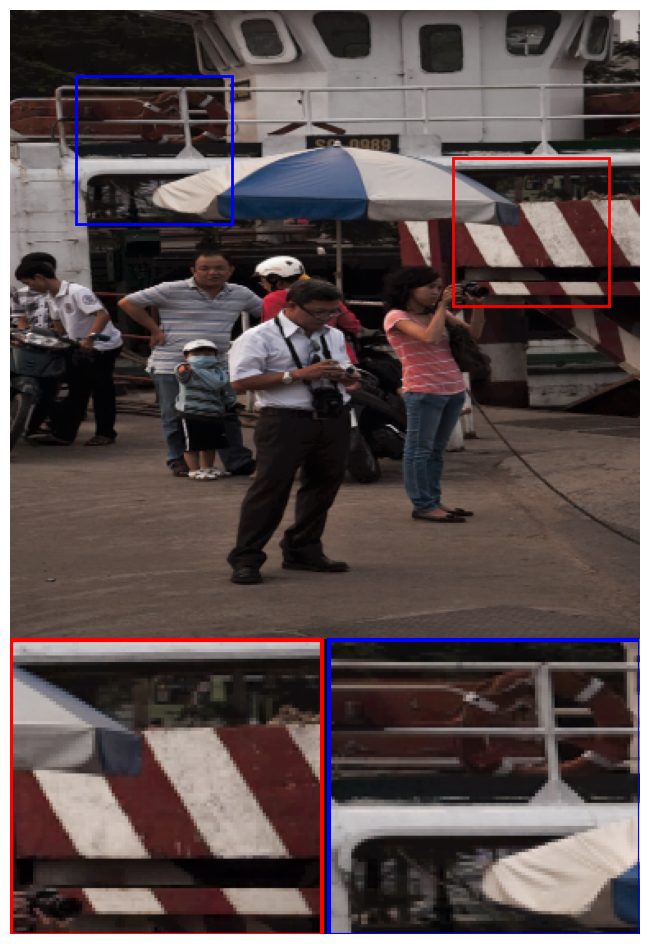} & 
		\includegraphics[width=0.115\textwidth, height=0.15\textheight]{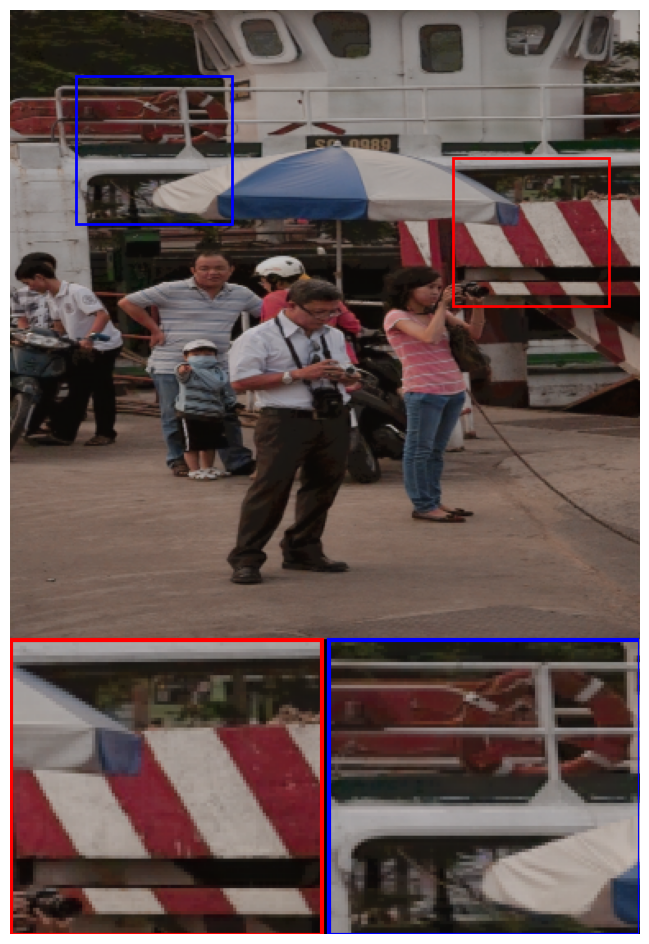} & 
		\includegraphics[width=0.115\textwidth, height=0.15\textheight]{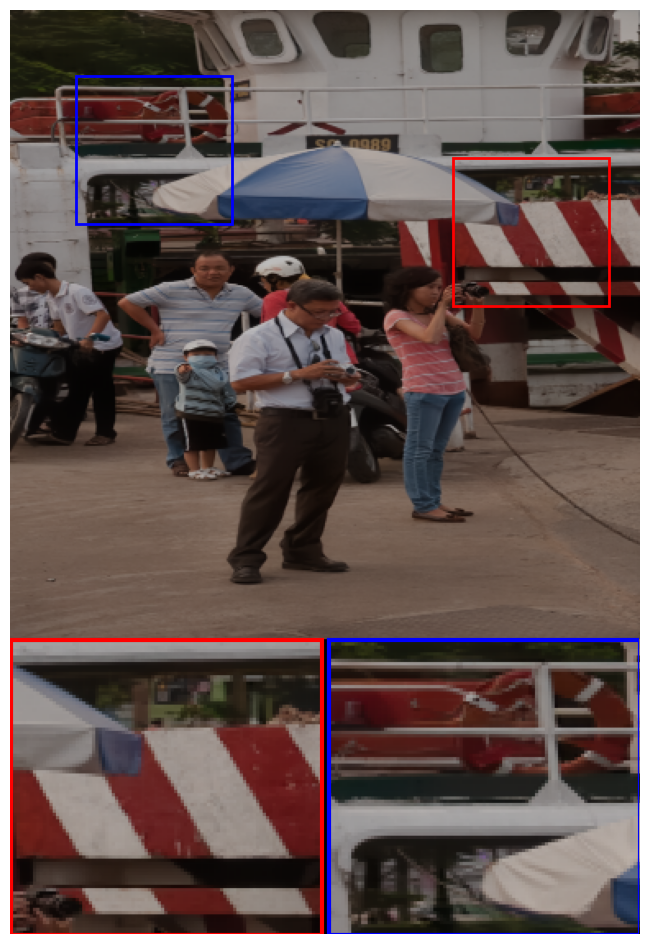} & 
		\includegraphics[width=0.115\textwidth, height=0.15\textheight]{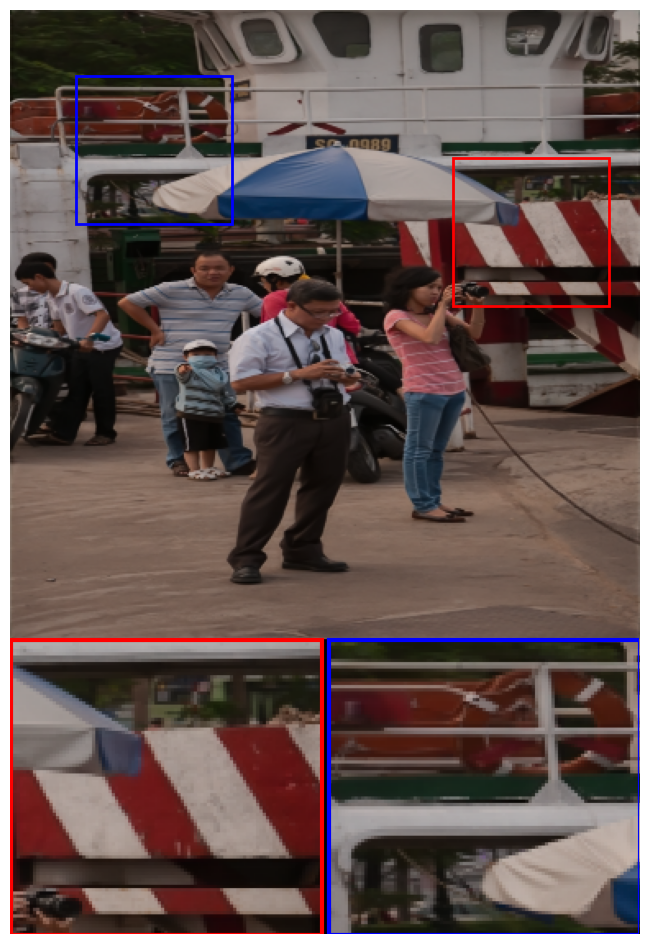} & 
		\includegraphics[width=0.115\textwidth, height=0.15\textheight]{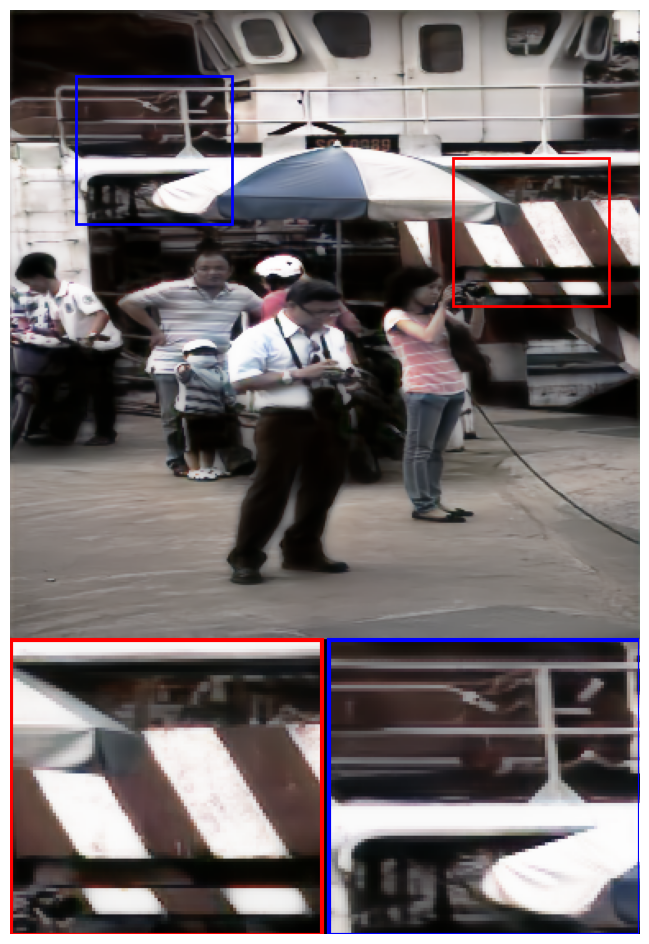} & 
		\includegraphics[width=0.115\textwidth, height=0.15\textheight]{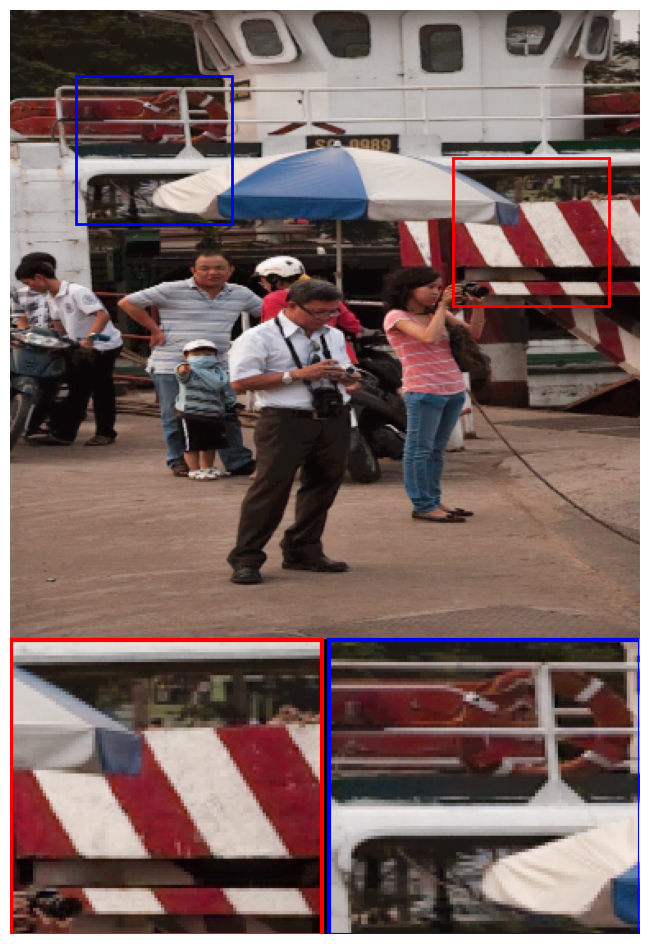} & 
		\includegraphics[width=0.115\textwidth, height=0.15\textheight]{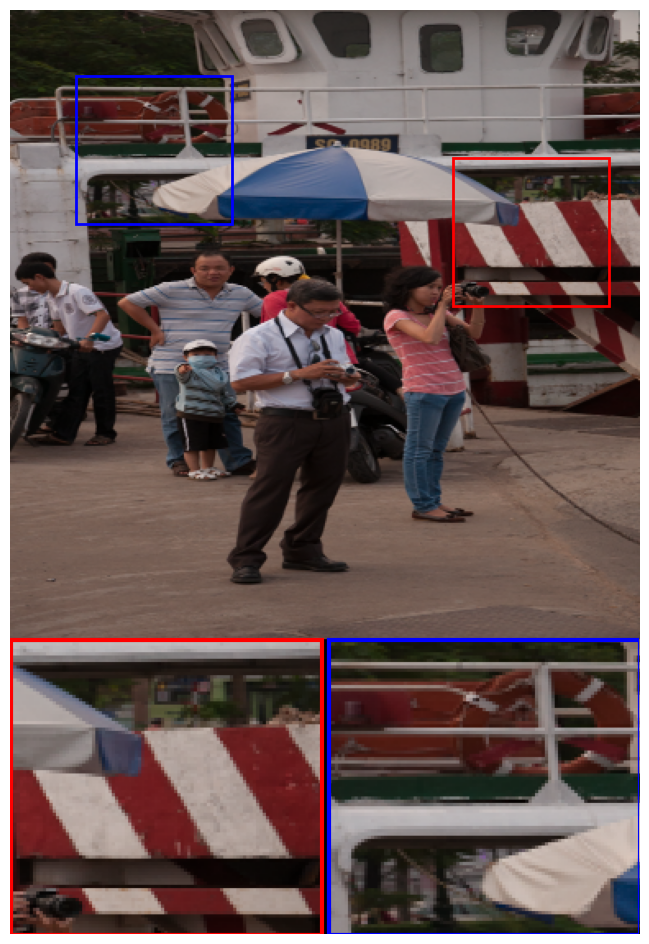} \\
		\includegraphics[width=0.115\textwidth, height=0.15\textheight]{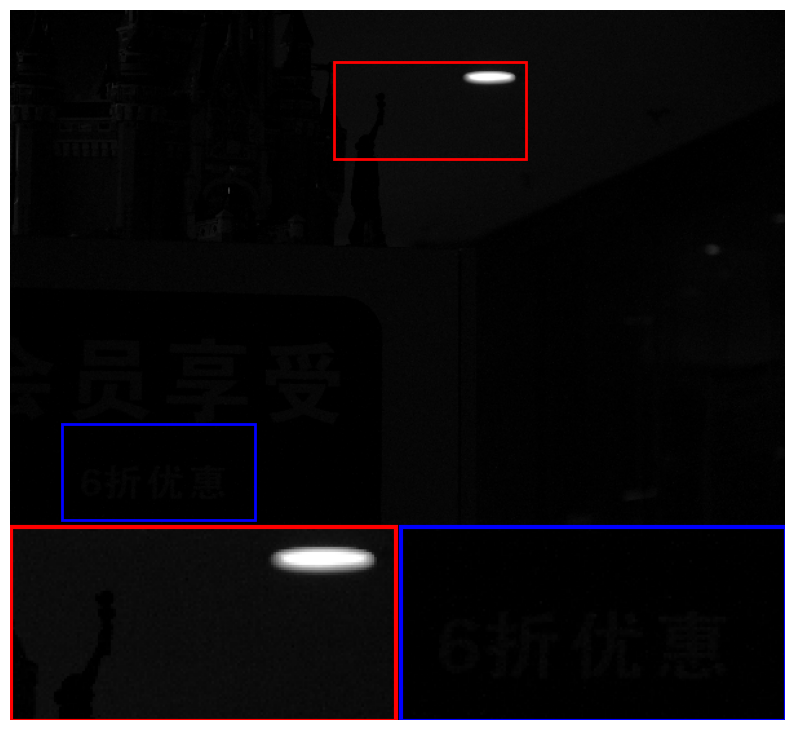} & 
		\includegraphics[width=0.115\textwidth, height=0.15\textheight]{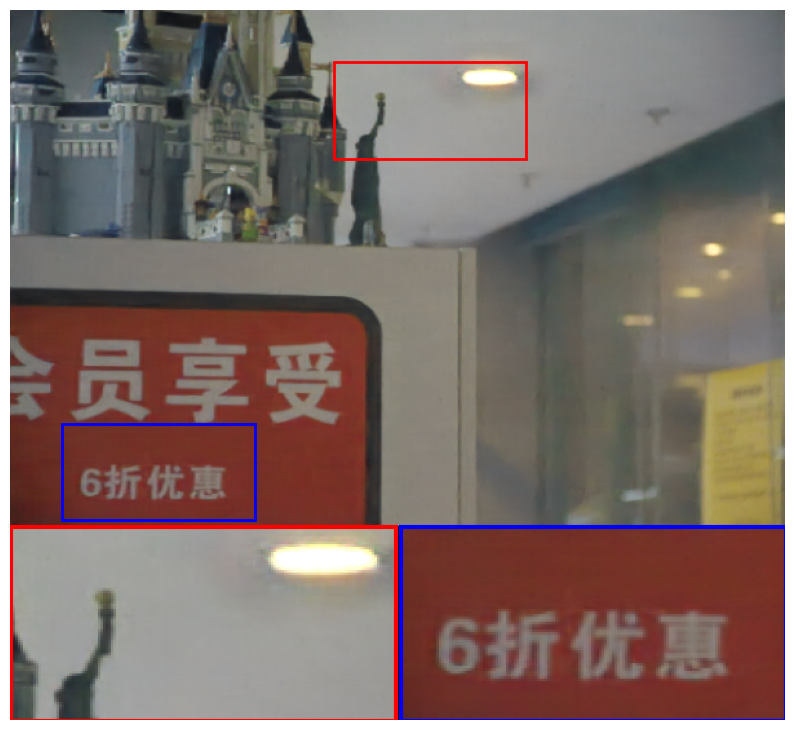} & 
		\includegraphics[width=0.115\textwidth, height=0.15\textheight]{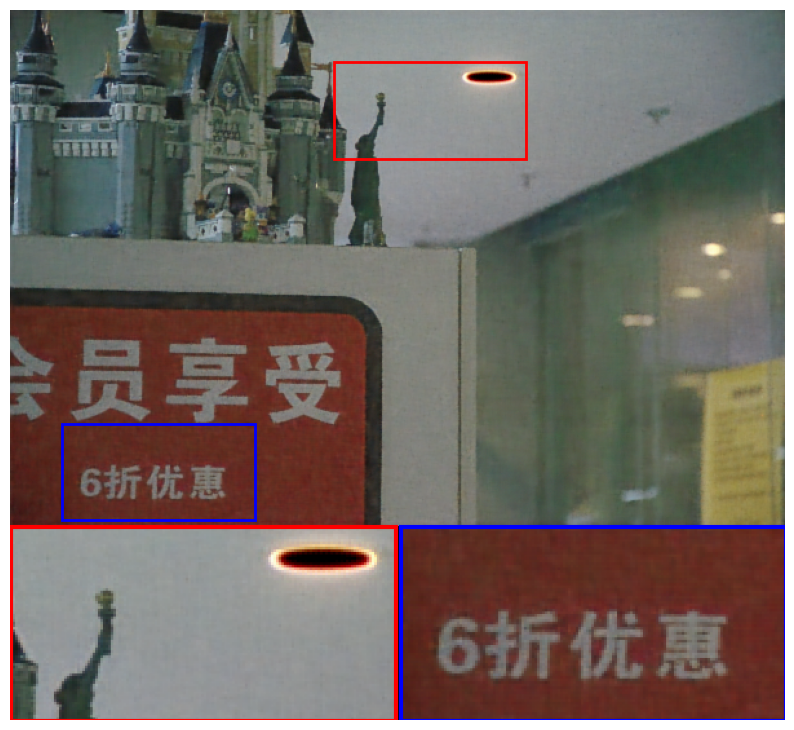} & 
		\includegraphics[width=0.115\textwidth, height=0.15\textheight]{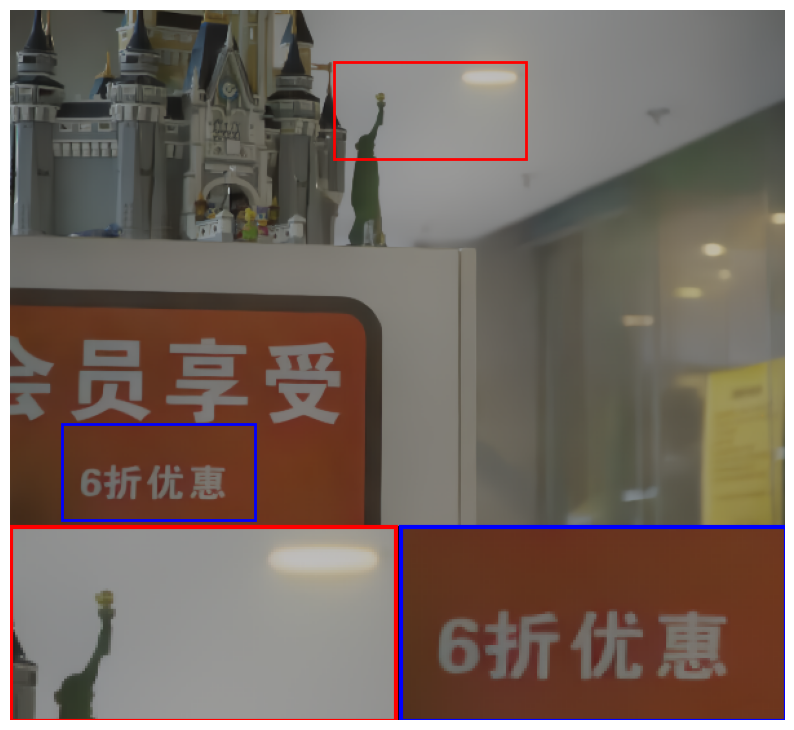} & 
		\includegraphics[width=0.115\textwidth, height=0.15\textheight]{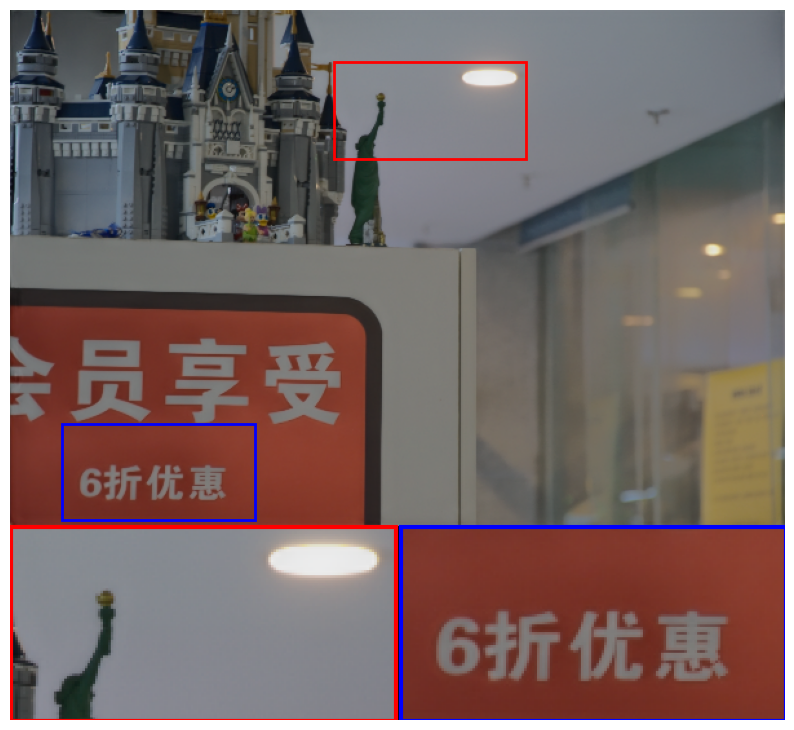} & 
		\includegraphics[width=0.115\textwidth, height=0.15\textheight]{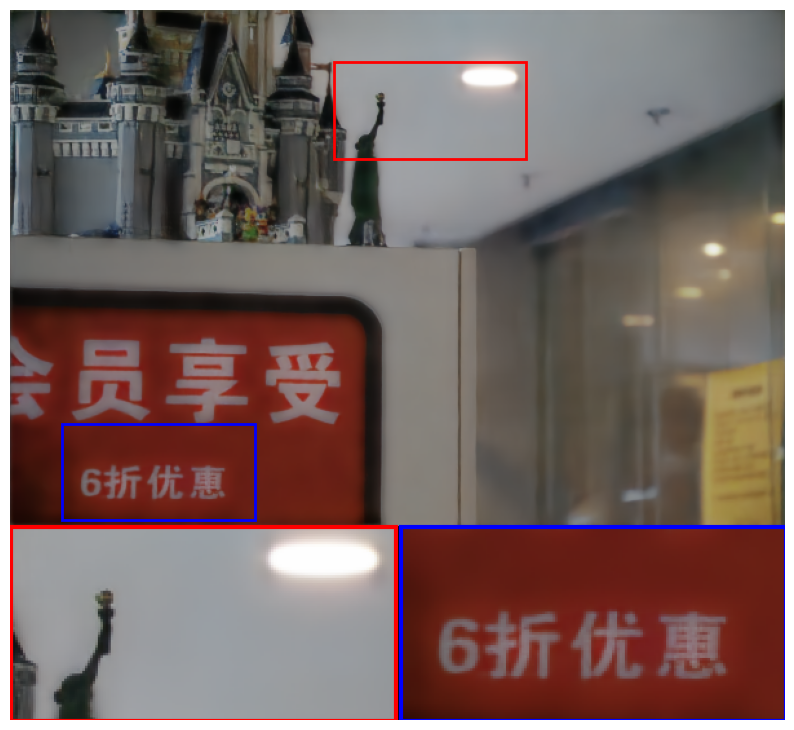} & 
		\includegraphics[width=0.115\textwidth, height=0.15\textheight]{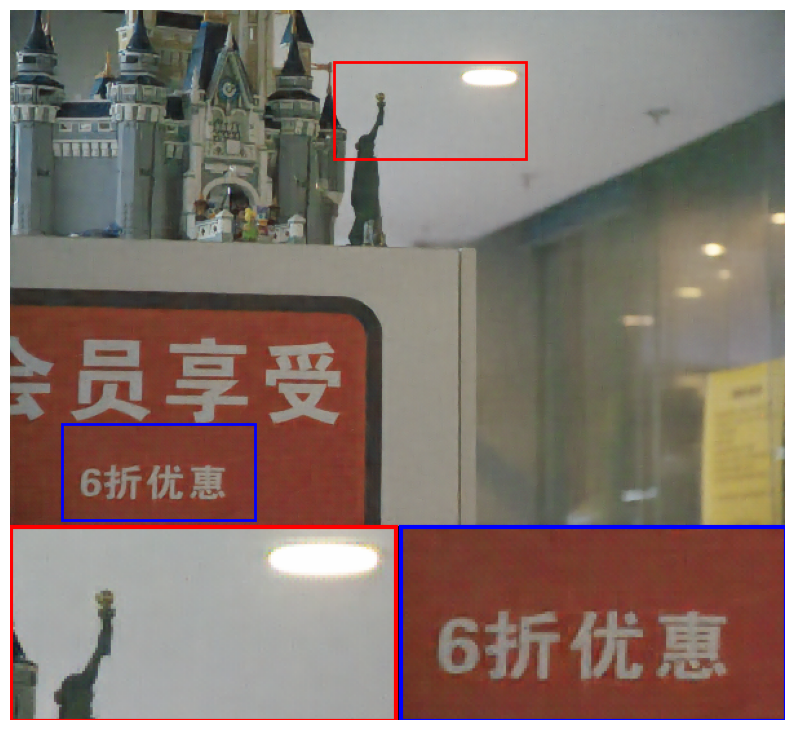} & 
		\includegraphics[width=0.115\textwidth, height=0.15\textheight]{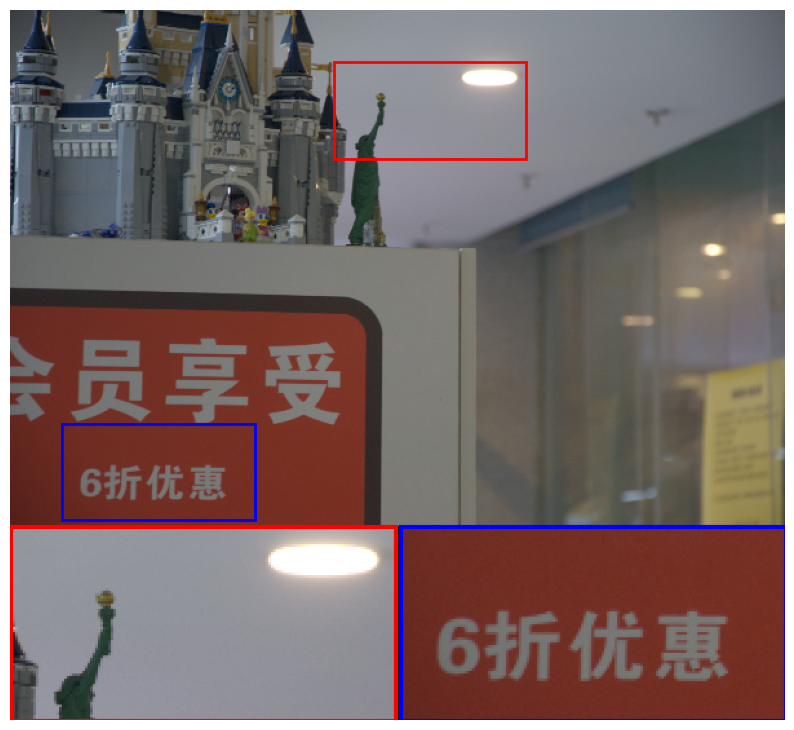} \\
		{\bfseries \small Input} & {\bfseries  \small SNR} \cite{exp5} & {\bfseries \small Retfomer} \cite{LLIE_2} & {\bfseries \small LLFlow} \cite{exp8} & {\bfseries \small Bread} \cite{LLIE_5} & {\bfseries \small KinD} \cite{exp6} & {\bfseries \small Ours} & {\bfseries \small GT} \\
	\end{tabular}
	\caption{Enhanced visualization comparison images generated by different state-of-the-art (SOTA) methods on LOLv2-Real (top row) and LOLv2-Synthetic (bottom row).}
	\label{fig:fig4}
\end{figure*}

\noindent of Fourier frequency-domain information. The primary function of this loss is to regularize the discrete points in the spectral map by projecting them into a Gaussian distribution. By calculating the distance between the Gaussian distributions of normal and low-light images, the model is guided to learn the Fourier frequency-domain information more effectively. The following content will elaborate on the specific implementation of this approach.

To measure the difference between the Fourier frequency-domain information of low-light and normal-light images from a distribution perspective, we consider using the Kullback-Leibler (KL) divergence. The KL-Divergence quantifies the difference between two data distributions from the perspective of relative entropy. Therefore, we use the KL-Divergence to measure the discrepancy between the predicted normal-light distribution and the true normal-light distribution. Let the predicted image distribution be denoted as $\pi_{\text{pred}}$ and the true image distribution as $\pi_{\text{true}}$. The formula for calculating the KL-Divergence between these two distributions is as follows:

\begin{equation}
	\label{KL1}
	\small 
	D_{KL}(\pi_{\text{pred}}(x) \parallel \pi_{\text{true}}(x)) = \int \pi_{\text{pred}}(x) \log \left( \frac{\pi_{\text{pred}}(x)}{\pi_{\text{true}}(x)} \right) dx
\end{equation}

This formula measures the discrepancy between the distribution of predicted normal-light images and that of the true normal-light images. This enables the model to reduce the discrepancy between these distributions during training, thereby approximating the Fourier frequency-domain distribution of normal-light images. It should be noted that calculating the divergence between the Fourier frequency-domain information of different images requires knowledge of their respective distributions. However, obtaining the exact distribution of a dataset is often challenging. To simplify the problem and enhance computational efficiency, we assume that the amplitude and phase components each follow a one-dimensional Gaussian distribution. By calculating the KL-Divergence between two Gaussian distributions, we can achieve distribution fitting.

To project the amplitude and phase components of the frequency spectrum into a Gaussian distribution, we need to determine their corresponding means and variances. These parameters are derived from calculations performed on the frequency spectrum channels. Initially, the image is transformed into its frequency spectrum using the Fast Fourier Transform (FFT). From the resulting frequency spectrum, the amplitude and phase components are extracted as shown in Fig. \ref{fig:fig1}. The mean and variance of these components are then calculated to obtain the necessary parameters for the Gaussian distribution. To ensure numerical stability, a small constant $1 \times 10^{-8}$ is added to the variance to prevent it from becoming zero. Finally, using these means and variances, the amplitude and phase maps are projected into a Gaussian distribution. The difference between two such Gaussian distributions can then be computed , as shown in Eq. (\ref{KL2}) and Eq. (\ref{KL3}). 

{\small
\begin{multline}
	\label{KL2}
	D_{KL-amp}\left( N(\mu_{\text{Pred-amp}}, \sigma_{\text{Pred-amp}}^2) \parallel N(\mu_{\text{True-amp}}, \sigma_{\text{True-amp}}^2) \right) \\
	= \log \left( \frac{\sigma_{\text{True-amp}}}{\sigma_{\text{Pred-amp}}} \right) + \frac{\sigma_{\text{Pred-amp}}^2 + (\mu_{\text{Pred-amp}} - \mu_{\text{True-amp}})^2}{2\sigma_{\text{True-amp}}^2} - \frac{1}{2}
\end{multline}
}

{\small
\begin{multline}
	\label{KL3}
	\small
	D_{KL-pha}\left( N(\mu_{\text{Pred-pha}}, \sigma_{\text{Pred-pha}}^2) \parallel N(\mu_{\text{True-pha}}, \sigma_{\text{True-pha}}^2) \right) \\
	= \log \left( \frac{\sigma_{\text{True-pha}}}{\sigma_{\text{Pred-pha}}} \right) + \frac{\sigma_{\text{Pred-pha}}^2 + (\mu_{\text{Pred-pha}} - \mu_{\text{True-pha}})^2}{2\sigma_{\text{True-pha}}^2} - \frac{1}{2}
\end{multline}
}

So, \( \mu_{\text{Pred-amp}} \) and \( \mu_{\text{True-amp}} \) denote the mean values of the predicted and true amplitudes, respectively, while \( \sigma_{\text{Pred-amp}} \) and \( \sigma_{\text{True-amp}} \) represent their variances. Similarly, \( \mu_{\text{Pred-pha}} \) and \( \mu_{\text{True-pha}} \) correspond to the mean values of the predicted and true phases, and \( \sigma_{\text{Pred-pha}} \) and \( \sigma_{\text{True-pha}} \) signify their variances. Furthermore, it is evident from the following discussion that this loss function is differentiable. In generative models, the KL-Divergence is frequently used as a loss function to measure the difference between distributions and it is derivable, which implies its differentiability \cite{IG_3, idea3}. Moreover, the Gaussian distribution is a smooth function with respect to its parameters (mean and variance) and thus differentiable. Consequently, the KL-Divergence between two Gaussian distributions is differentiable with respect to their parameters, which ensures that it can be effectively used for gradient-based optimization in neural networks.

By reducing the complexity of the frequency-domain information to the parameters of a one-dimensional Gaussian distribution and computing the divergence between two Gaussian distributions, we accomplish the fitting of amplitude and phase characteristics from a distributional standpoint. The divergence between these Gaussian distributions is utilized as the \( L_{FKL} \) loss, as depicted in Eq. (\ref{LFKL}).

\begin{equation}
	\label{LFKL}
	L_{FKL} = \frac{(D_{KL-amp} + D_{KL-pha})}{2}
\end{equation}

\subsection{VGG With KL-Divergence}
To better leverage the feature extraction capabilities of the pre-trained VGG loss, we propose an improvement to the pre-trained VGG network by incorporating the KL-Divergence. The specific approach is as follows:

Both the input ground-truth image and the predicted image are fed into the pre-trained deep VGG network for feature extraction. The extracted features of the ground-truth and predicted images are then flattened into two-dimensional arrays of size $[B, C \times H \times W]$, where $B$ denotes the batch size, $C$ is the number of channels, and $H \times W$ represents the spatial dimensions of the feature maps.

These flattened feature arrays of the ground-truth and predicted images are treated as discrete distributions. The difference between these distributions is quantified using the discrete KL-Divergence formula, which is given as follows Eq. (\ref{KL4}):

{\small
\begin{equation}
	\label{KL4}
	L_{VggKL} = D_{\text{KL}}(\pi_{\text{Pred}} \parallel \pi_{\text{True}}) = \sum \pi_{\text{Pred}}(x) \log \left( \frac{\pi_{\text{Pred}}(x)}{\pi_{\text{True}}(x)} \right)
\end{equation}
}

The KL-Divergence between the predicted feature array and the ground-truth feature array is computed and returned as the loss. By improving the pre-trained VGG network with KL-Divergence, the model's ability to extract features is further enhanced, as will be seen in the subsequent experiments.

\subsection{Combined Losses}
In low-light imaging tasks, the use of a composite loss function for model training is highly prevalent \cite{LLIE_4, fourier_LLapp3, fourier_LLapp4}. Composite losses integrate the advantages of multiple individual loss functions, thereby enabling a more comprehensive evaluation of model performance. This approach helps enhance the generalization capability and robustness of the model. The composite loss is formulated as shown in Eq. (\ref{cl2}), where the parameters are carefully tuned to balance the contributions of the different loss functions.

{\small
\begin{multline}
	\label{cl2}
	L = (\alpha_1 L_S + \alpha_2 L_{Hist} + \alpha_3 L_{MS-SSIM} +  \alpha_4 L_{PSNR} + \\ \alpha_5 L_{Color})_{\text{Base}} + \alpha_6 L_{VggKL} + \alpha_7 L_{FKL}	
\end{multline}
}

The Smooth L1 loss $L_S$ ensures accuracy while providing robustness against outliers. The histogram loss $L_{\text{Hist}}$ maintains similarity in brightness distribution. The multi-scale SSIM loss $L_{\text{MS-SSIM}}$ preserves structural similarity across different scales. The PSNR loss $L_{\text{PSNR}}$ optimizes the peak signal-to-noise ratio to enhance objective quality metrics. The color loss $L_{\text{Color}}$ prevents color bias and ensures color balance. These five losses---$L_S$, $L_{\text{Hist}}$, $L_{\text{MS-SSIM}}$, $L_{\text{PSNR}}$, and $L_{\text{Color}}$---constitute the base loss. $L_{VggKL}$  is the improved VGG perceptual loss, which leverages KL-Divergence to ensure consistency in the distribution of extracted features. $L_{FKL}$ is the Fourier loss improved by Gaussian KL-Divergence, which compares the distributions of amplitude and phase information obtained after fast fourier transformation.

The corresponding weights for these losses are set as $(a_1, a_2, a_3, a_4, a_5, a_6, a_7)$ = $(1, 0.06, 0.05, 0.5, 0.0083, 0.15, \\ 0.1)$. The weights $(a_1, a_2, a_3, a_4, a_5, a_6)$ are empirically determined, as shown in \cite{LLIE_4}. These losses are flexibly combined through adjustable weighting coefficients to comprehensively optimize the quality of low-light image enhancement. The weight $a_7$, which controls the proportion of the Fourier KL-Divergence loss, is determined through ablation studies to be optimal at around $0.1$, at can be seen in Figure \ref{fig:loss_rate}.

\section{Experiments And Analysis}
This section presents the datasets, implementation details used in our experiments. We then conduct comparison experiments from both qualitative and quantitative perspectives to demonstrate the superiority of our approach compared with state-of-the-art methods. Finally, we perform ablation studies to verify the effectiveness of the proposed modules, loss functions and Hyperparameter of Fourier-KL Loss.

\begin{figure*}[h]
	\includegraphics[height=0.35\textheight,width=\linewidth]{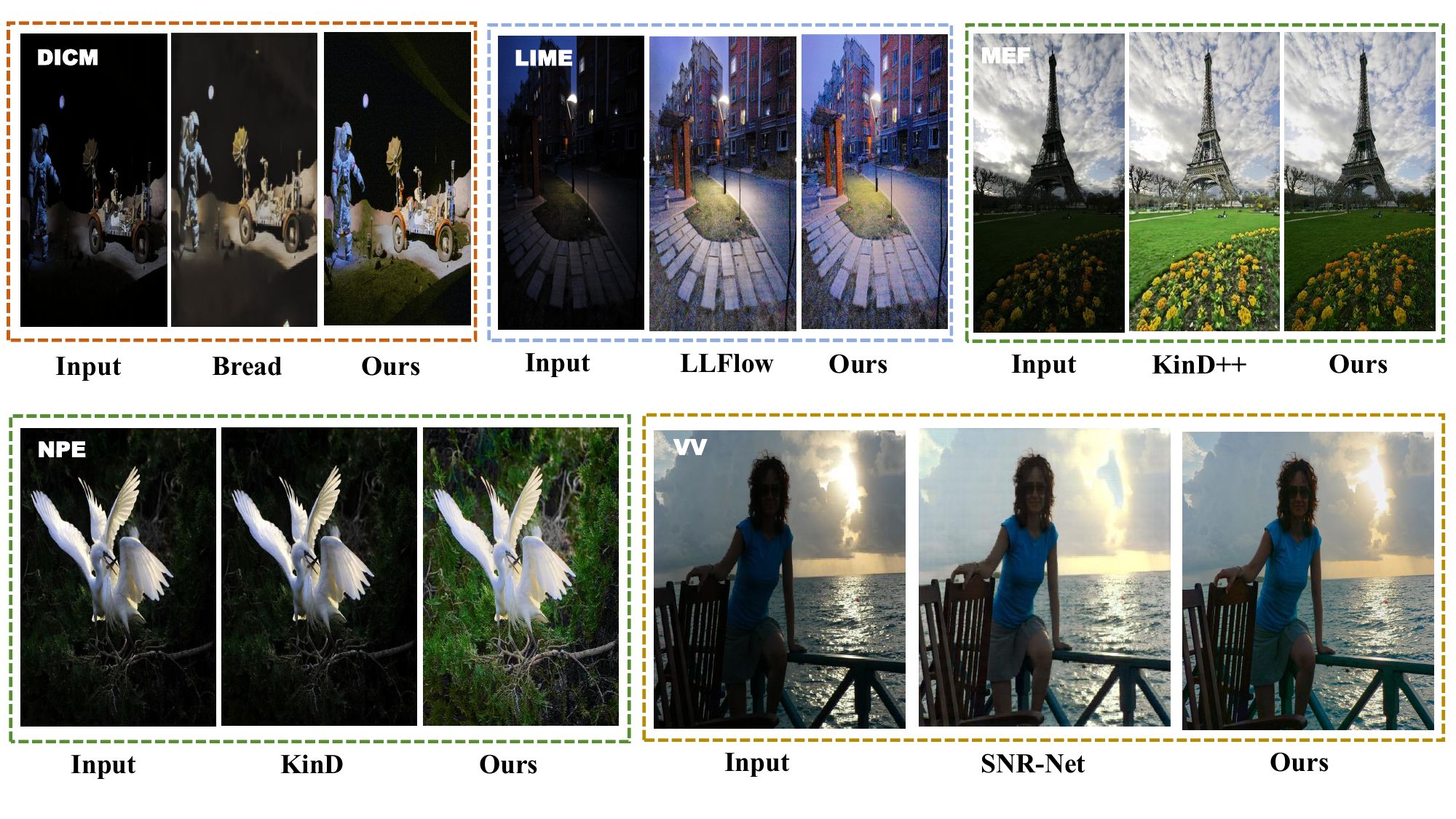}
	\caption{Qualitative comparison results on the LIME, VV, DICM, NPE, and MEF datasets. We selected one image from each dataset to compare our method with other models.}
	\label{fig:fig5}
\end{figure*}

\begin{table*}[h]
	\centering
	\caption{Quantitative results of PSNR/SSIM/LPIPS on the LOL (v1 and v2) datasets. FLOPS tested on a single 256x256 image. Best performance highlighted in red, and second-best performance in cyan.}
	\label{tab:tb1}
	\renewcommand{\arraystretch}{1.5}
	\setlength{\tabcolsep}{8pt}
	\begin{tabular}{c|cc|ccc|ccc|ccc}
		\hline
		\multicolumn{1}{c|}{\multirow{2}{*}{\centering \bfseries Methods}} &
		\multicolumn{2}{c|}{\bfseries Complexity} &
		\multicolumn{3}{c|}{\bfseries LOL-v1} &
		\multicolumn{3}{c|}{\bfseries LOL-v2-Real} &
		\multicolumn{3}{c}{\bfseries LOL-v2-Synthetic} \\
		
		& FLOPS(G) & Parama(M) & PSNR & SSIM & LPIPS & PSNR & SSIM & LPIPS & PSNR & SSIM & LPIPS \\
		\hline
		Bread \cite{LLIE_5} & 7.57 & 2.02 & 25.29 & 0.847 & 0.155 & 20.83 & 0.847 & 0.174 & 17.62 & 0.919 & 0.091 \\
		PairLIE \cite{exp1} & 20.81 & 0.33 & 23.52 & 0.755 & 0.248 & 19.88 & 0.778 & 0.317 & 19.07 & 0.794 & 0.230 \\
		EnGAN \cite{LLIE_8} & 61.01 & 114.35 & 20.00 & 0.691 & 0.317 & 18.23 & 0.617 & 0.309 & 16.57 & 0.734 & 0.220 \\
		LLFlow \cite{exp8} & 358.4 & 17.42 & 24.99 & \textcolor{red}{0.871} & \textcolor{cyan}{0.117} & 17.43 & 0.831 & 0.176 & 24.81 & 0.919 & 0.067 \\
		Restormer \cite{exp3} & 144.25 & 26.13 & 26.68 & \textcolor{cyan}{0.853} & 0.147 & 26.12 & 0.853 & 0.232 & 25.43 & 0.859 & 0.144 \\
		LLFormer \cite{LLIE_7} & 22.52 & 24.55 & 25.76 & 0.823 & 0.167 & 20.06 & 0.792 & 0.211 & 24.03 & 0.909 & 0.066 \\
		RetinexMamba \cite{exp11} & - & - & 24.03 & 0.827 & 0.146 & 22.45 & 0.844 & 0.174 & 25.89 & 0.935 & \textcolor{cyan}{0.054} \\
		RetinexNet \cite{LLIE_1} & 584.47 & 0.84 & 18.92 & 0.427 & 0.470 & 18.32 & 0.447 & 0.543 & 19.09 & 0.774 & 0.255 \\
		SNR-Aware \cite{exp5} & 26.35 & 4.01 & \textcolor{cyan}{26.72} & 0.851 & 0.152 & 27.21 & \textcolor{cyan}{0.871} & \textcolor{cyan}{0.163} & 27.79 & \textcolor{cyan}{0.941} & 0.056 \\
		KinD \cite{exp6} & 34.99 & 8.02 & 23.01 & 0.843 & 0.156 & 20.01 & 0.842 & 0.375 & 22.62 & 0.904 & 0.252 \\
		QuadPrior \cite{exp10} & 1103.20 & 1252.75 & 22.85 & 0.800 & 0.201 & 20.59 & 0.811 & 0.202 & 16.11 & 0.758 & 0.114 \\
		Retinexformer \cite{LLIE_2} & 15.85 & 1.53 & 26.52 & 0.846 & 0.129 & \textcolor{cyan}{27.70} & 0.856 & 0.171 & \textcolor{cyan}{29.03} & 0.938 & 0.059 \\
		Ours & 10.93 & 0.923 & \textcolor{red}{26.77} & 0.843 & \textcolor{red}{0.101} & \textcolor{red}{28.63} & \textcolor{red}{0.883} & \textcolor{red}{0.107} & \textcolor{red}{29.39} & \textcolor{red}{0.944} & \textcolor{red}{0.044} \\
		\hline
	\end{tabular}
\end{table*}

\subsection{Datasets}
We conducted both qualitative and quantitative evaluations of our method on the public paired low-light image datasets LOL and LSRW-HUAWEI, and performed qualitative tests on 

\begin{table}[h]
	\centering
	\caption{Quantitative results on the LSRW-Huawei dataset. The best performance is indicated in bold.}
	\label{tab:tb2}
	\scriptsize 
	\renewcommand{\arraystretch}{1.5} 
	\setlength{\tabcolsep}{1.6pt} 
	\begin{tabular}{lcccccc}
		\toprule
		{\bfseries Methods} & {\bfseries LIME \cite{datasets_LIME}} & {\bfseries KinD \cite{exp6}} & {\bfseries KinD++ \cite{exp7}} & {\bfseries SNR-Aware \cite{exp5}} & {\bfseries Bread \cite{LLIE_5}} \\
		\midrule
		PSNR & 17.00 & 16.58 & 15.43 & 20.67 & 19.20 \\
		SSIM & 0.3816 & 0.5690 & 0.5695 & 0.5910 & 0.6179 \\
		\midrule
		{\bfseries Methods} & {\bfseries RetFormer \cite{LLIE_2}} & {\bfseries MIRNet \cite{exp2}} & {\bfseries Four \cite{fourier_LLapp2}} & {\bfseries DMFour \cite{fourier_LLapp4}} & {\bfseries Ours} \\
		\midrule
		PSNR & 21.23 & 19.98 & 21.30 & 21.47 & \textbf{23.17} \\
		SSIM & 0.6309 & 0.6085 & 0.6220 & 0.6331 & \textbf{0.6332} \\
		\bottomrule
	\end{tabular}
\end{table}

\begin{table}[h]
	\centering
	\caption{Quantitative results NIQE on the five unpaired datasets. The best performance is highlighted in red, and the second-best performance is indicated in cyan. All evaluation methods used weights pretrained on the LOL-v2-Synthetic dataset.}
	\label{tab:tb3}
	\setlength{\tabcolsep}{6pt} 
	\begin{tabular}{ccccccc}
		\toprule
		{\bfseries Methods} & {\bfseries LIME } & {\bfseries VV } & {\bfseries DICM } & {\bfseries NPE} & {\bfseries MEF} & {\bfseries AVG} \\
		\midrule
		Kind & 4.772 & 3.835 & \textcolor{cyan}{3.614} & 4.175 & 4.819 & 4.194 \\
		MIRNet \cite{exp2} & 6.453 & 4.735 & 4.042 & 5.235 & 5.504 & 5.101 \\
		Sparse \cite{exp4} & 5.451 & 4.884 & 4.733 & 5.208 & 5.754 & 5.279 \\
		FECNet \cite{fourier_app6} & 6.041 & 3.346 & 4.139 & 4.500 & 4.707 & 4.336 \\
		HDMNet \cite{exp9} & 6.403 & 4.462 & 4.773 & 5.108 & 5.993 & 5.056 \\
		Bread & 4.717 & 3.304 & 4.179 & 4.160 & 5.369 & 4.194 \\
		Retinexformer & \textcolor{red}{3.441} & 3.706 & 4.008 & \textcolor{red}{3.893} & \textcolor{cyan}{3.727} & \textcolor{cyan}{3.755} \\
		FourLLIE & 4.402 & \textcolor{cyan}{3.168} &\textcolor{red}{3.374} & \textcolor{cyan}{3.909} & 4.362 & 3.907 \\
		Ours & \textcolor{cyan}{4.034} & \textcolor{red}{3.101} & 3.933 & 3.983 & \textcolor{red}{3.426} & \textcolor{red}{3.695} \\
		\bottomrule
	\end{tabular}
\end{table}

\noindent five additional unpaired datasets. The details of these datasets are as follows:

\textbf{LOL.} The LOL dataset comprises two versions: v1 \cite{LLIE_1} and v2 \cite{exp4}. LOL-v1 is a mixed dataset containing both real and synthetic images, with training and testing splits of 485:15 (training/testing). LOL-v2 further separates real and synthetic data and introduces additional new datasets. The training and testing splits for LOL-v2-Real and LOL-v2-Synthetic are 689:100 and 900:100, respectively.

\textbf{LSRW-Huawei.} LSRW-Huawei \cite{datasets_LSRWHuawei} consists of real scenes captured using different devices, similar to the LOL-Real dataset. It contains 3150 training data pairs and 20 testing data pairs. 

\textbf{Unpaired DataSets.} We evaluated our method on the following five unpaired datasets: DICM \cite{datasets_DICM}, LIME \cite{datasets_LIME}, MEF \cite{datasets_MEF}, NPE \cite{datasets_NPE}, and VV \cite{datasets_VV}, which contain 64, 10, 17, 85, and 24 images, respectively.

\textbf{Table \ref{tab:tb1}.} It provides a detailed quantitative comparison of various methods on the LOL (v1 and v2) datasets, with a particular focus on PSNR and SSIM values, which are critical metrics for evaluating image enhancement quality. Our proposed method achieved the best performance under the same testing conditions. Specifically, on LOL-v2-Real and LOL-v2-Synthetic, our method achieved PSNR values of 28.63 and 29.39, and SSIM values of 0.883 and 0.944, respectively, significantly outperforming other existing methods. Additionally, our method demonstrated lower computational complexity (FLOPS) and parameter count (Param), with values of 10.93 G and 0.923M, respectively, indicating significant advantages in both efficiency and performance.

\textbf{Table \ref{tab:tb2}.} It focuses on the comparison of PSNR and SSIM values on another dataset. The PSNR and SSIM values achieved by our method are 23.17 and 0.6332, respectively, surpassing all other methods. This result further demonstrates the effectiveness and robustness of our image enhancement method under different conditions, especially when dealing with challenging image data.

\textbf{Table \ref{tab:tb3}.} It presents the NIQE scores of various image enhancement methods evaluated across five benchmark datasets (LIME, VV, DICM, NPE, and MEF), along with the corresponding average score (AVG) for each method. The proposed approach achieves the lowest NIQE score on all five datasets, yielding an overall average of 3.695. This result demonstrates that our method consistently produces higher-quality enhanced images, characterized by reduced noise and superior perceptual quality. In contrast, competing approaches exhibit higher ave-rage scores, with Retinexformer achieving the second-lowest average of 3.755, while HDMNet records the highest average of 5.056.

\begin{table}[h]
	\centering
	\caption{Testing different combinations of loss functions and their effects on the LOL (v1 and v2) datasets. The best performance is indicated in bold.}
	\scriptsize 
	\label{tab:loss}
	\begin{tabular}{c|cc|cc|cc}
		\toprule
		\multirow{2}{*}{\bfseries Methods} & \multicolumn{2}{c|}{\bfseries LOL-V1} & \multicolumn{2}{c|}{\bfseries LOL-V2-Real} & \multicolumn{2}{c}{\bfseries LOL-V2-Synthetic} \\
		& PSNR & SSIM & PSNR & SSIM & PSNR & SSIM \\
		\hline
		Base & 25.74 & 0.807 & 27.33 & 0.853 & 28.55 & 0.936 \\
		Base+$L_{F}$ & 26.12 & 0.823 & 27.27 & 0.850 & 28.83 & 0.939 \\
		Base+$L_{FKL}$ & 26.16 & 0.822 & 27.57 & 0.856 & 29.06 & 0.940 \\
		Base+$L_{Vgg}$ & 26.13 & 0.834 & 28.39 & 0.875 & 28.54 & 0.937 \\
		Base+$L_{VggKL}$ & 26.20 & 0.835 & 28.56 & 0.882 & 29.18 & 0.942 \\
		Full & \textbf{26.77} & \textbf{0.843} & \textbf{28.63} & \textbf{0.884} & \textbf{29.39} & \textbf{0.944} \\
		\bottomrule
	\end{tabular}
\end{table}

\subsection{Ablation Study}
\textbf{Different loss functions.} The Table \ref{tab:loss} presents a comparison of PSNR and SSIM values on the LOL-V1, LOL-V2-Real,

\begin{figure}[!t]
	\centering
	\begin{tabular}{ccc}
		\includegraphics[width=0.28\linewidth,height=1in]{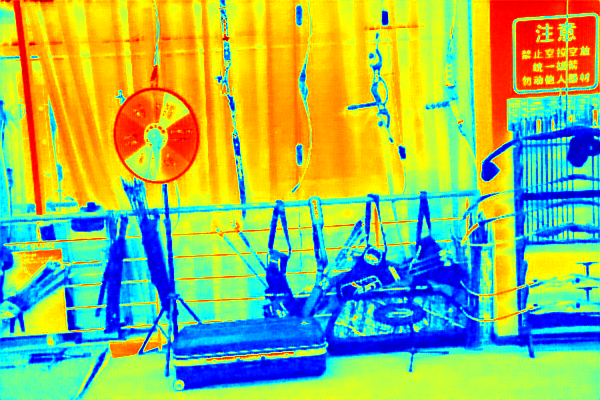} & 
		\includegraphics[width=0.28\linewidth,height=1in]{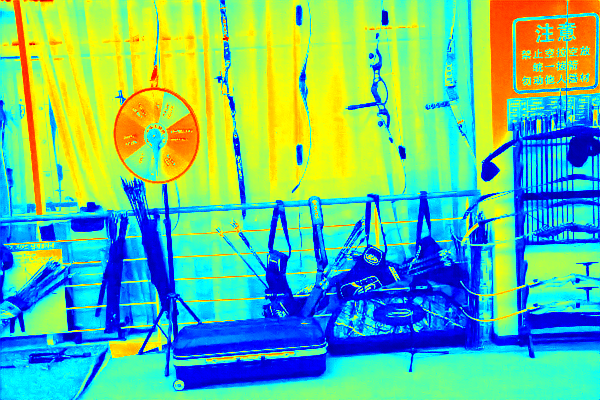} & 
		\includegraphics[width=0.28\linewidth,height=1in]{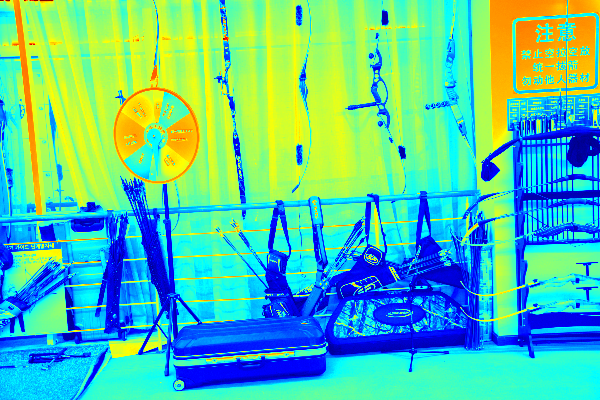} \\
		{\bfseries \small Base-CAM} & {\bfseries \small Full-CAM} & {\bfseries \small GT-CAM}
	\end{tabular}
	\caption{The Base-CAM represents the heatmap of the image obtained by training with the Base loss function. The Full-CAM denotes the heatmap derived from training with the Full loss function. In contrast, the GT-CAM corresponds to the heatmap generated based on the Ground Truth.}
	\label{fig:cam}
\end{figure}

\noindent and LOL-V2-Synthetic datasets. The "Base" loss serves as the foundation, other with variants incorporating various loss functions. Specifically, "Base+$L_{F}$" adds an MSE-based loss, as shown in \cite{fourier_app3, fourier_LLapp2, fourier_app5}, to fit Fourier frequency information; "Base+$L_{FKL}$" replaces the MSE with a KL-Divergence loss for the same purpose; "Base+$L_{Vgg}$" incorporates a pre-trained VGG network as a perceptual loss \cite{LLIE_4}; "Base+$L_{VggKL}$" combines VGG with KL-Divergence as the perceptual loss and "Full" represents the complete loss function comprising "Base + $L_{FKL}$ + $L_{VggKL}$". In the ablation studies of different low-light image enhancement methods, the tabulated data demonstrate that the loss functions improved by introducing the Kullback-Leibler (KL) divergence exhibit significant effectiveness.

As shown in Figure \ref{fig:cam}, the Full-CAM, which is the heatmap resulting from training with the Full loss function, closely mirrors the GT-CAM. This similarity indicates that the Full loss function effectively captures the regions of interest that align with the ground truth activations, as opposed to the Base-CAM which is derived from training with the Base loss function.

\textbf{Different Modules.} To validate the proposed network modules, we conducted ablation experiments to assess their impact on image enhancement performance. We sequentially removed the CAB, IEL, SE, and DANCE modules from the full model and compared these variants with the complete model. Results showed that removing the CAB module led to a PSNR and 

\begin{table}[h]
	\centering
	\caption{Testing the effects of different modules on the LOLv1 dataset using the control variable method. The best performance is indicated in bold.}
	\label{tab:ablation1}
	\begin{tabular}{cccccc}
		\toprule
		CAB & IEL & SE & DANCE & PSNR & SSIM \\
		\midrule
		& $\checkmark$ & $\checkmark$ & $\checkmark$ & 26.23 & 0.833 \\
		$\checkmark$ &  & $\checkmark$ & $\checkmark$ & 26.06 & 0.832 \\
		$\checkmark$ & $\checkmark$ &  & $\checkmark$ & 26.05 & 0.830 \\
		$\checkmark$ & $\checkmark$ & $\checkmark$ &  & 25.97 & 0.828 \\
		$\checkmark$ & $\checkmark$ & $\checkmark$ & $\checkmark$ & \textbf{26.77} & \textbf{0.843} \\
		\bottomrule
	\end{tabular}
\end{table}

\begin{figure}[!t]
	\centering
	\includegraphics[width=3.5in,height=2.4in]{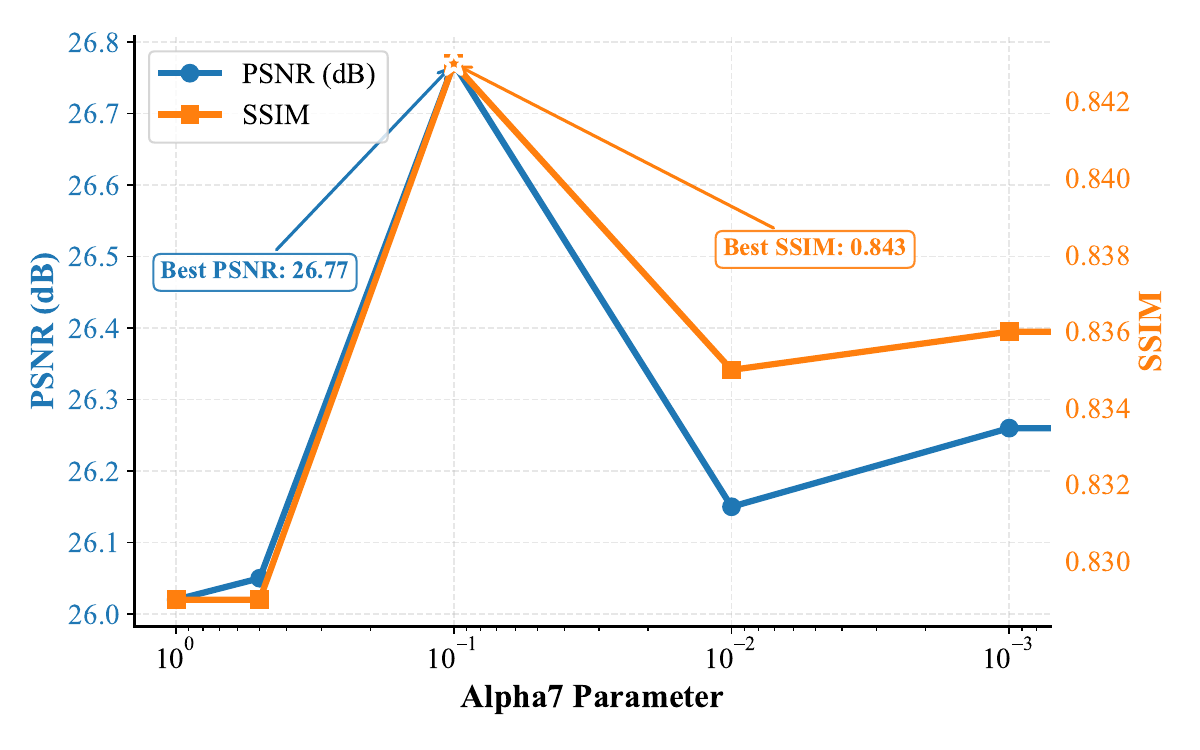}
	\caption{Testing the impact of different proportions of the loss function on model training results on the LOLv1 dataset.}
	\label{fig:loss_rate}
\end{figure}

\noindent SSIM decrease of 2.0\% and 1.1\%, respectively, indicating its significance in restoring image details. Removing the IEL module resulted in a PSNR decrease of 2.6\% and an SSIM decrease of 1.3\%, demonstrating its contribution to overall image quality. Removing the SE module also showed a downward trend in PSNR and SSIM, highlighting its importance in local feature enhancement. Removing the DANCE module resulted in the most noticeable decrease in PSNR and SSIM, with drops of 2.9\% and 2.0\%, respectively, underscoring its role in capturing dynamic features and texture details. In contrast, the full model achieved best PSNR and SSIM values, significantly outperforming other combinations. This demonstrates the synergistic effect of all modules in improving image enhancement accuracy and quality.

\textbf{Hyperparameter of Fourier-KL Loss.} The ablation experiment, as shown in Figure \ref{fig:loss_rate}, was primarily conducted on the LOLv1 dataset to evaluate the impact of different proportions of the $L_{FKL}$ loss function on image enhancement performance. The results indicated which is shown in Figure \ref{fig:loss_rate} that when the proportion of $L_{FKL}$ was set to 0.1, the model achieved the highest PSNR and SSIM values of 26.77 and 0.843, respectively, demonstrating optimal image enhancement performance at this ratio. Conversely, when the $L_{FKL}$ proportion was set to 1, 0.5, 0.01, 0.001, or when the $L_{FKL}$ loss function was omitted entirely, the model's PSNR and SSIM values were underperforming, as shown in Table \ref{tab:loss}. These values were all lower than those obtained at the optimal ratio of 0.1. This suggests that the $L_{FKL}$ loss function can significantly enhance model performance when the proportion is appropriately set, whereas high or low proportions lead to a decline in performance. Therefore, the proper setting of the $L_{FKL}$ proportion is crucial for achieving effective image enhancement, and a ratio of 0.1 around is identified as an optimal proportion for the $L_{FKL}$ loss function in this experiment.

\section{Conclusions}
Inspired by prior work on Fourier-based low-light image enhancement, this paper proposes a novel network architecture, LLFDisc, which integrates cross-attention and gating into a U-shaped image enhancement network to enhance the performance of low-light image enhancement in the Fourier frequency domain.We introduce a distribution-aware Fourier loss that models amplitude–phase statistics as Gaussian distributions and minimizes their divergence via a closed-form KL-Divergence objective, enabling robust global fitting without the local-bias of pixel-wise losses. Likewise, we upgrade the VGG perceptual loss by applying the same KL-Divergence principle to deep features, further improving structural fidelity. Ablation studies verify the individual contributions of each component. Extensive experiments on LOL-v1, LOL-v2 and LSRW-Huawei show that LLFDisc surpasses both Fourier and non-Fourier state-of-the-art methods in quantitative metrics and visual quality. 

\bibliographystyle{IEEEtran} 
\bibliography{IEEERef.bib}

\begin{thebibliography}{10}
\providecommand{\url}[1]{#1}
\csname url@samestyle\endcsname
\providecommand{\newblock}{\relax}
\providecommand{\bibinfo}[2]{#2}
\providecommand{\BIBentrySTDinterwordspacing}{\spaceskip=0pt\relax}
\providecommand{\BIBentryALTinterwordstretchfactor}{4}
\providecommand{\BIBentryALTinterwordspacing}{\spaceskip=\fontdimen2\font plus
\BIBentryALTinterwordstretchfactor\fontdimen3\font minus
  \fontdimen4\font\relax}
\providecommand{\BIBforeignlanguage}[2]{{%
\expandafter\ifx\csname l@#1\endcsname\relax
\typeout{** WARNING: IEEEtran.bst: No hyphenation pattern has been}%
\typeout{** loaded for the language `#1'. Using the pattern for}%
\typeout{** the default language instead.}%
\else
\language=\csname l@#1\endcsname
\fi
#2}}
\providecommand{\BIBdecl}{\relax}
\BIBdecl

\bibitem{fourier_LLapp2}
C.~Wang, H.~Wu, and Z.~Jin, ``Fourllie: Boosting low-light image enhancement by
  fourier frequency information,'' in \emph{Proceedings of the 31st ACM
  International Conference on Multimedia}, 2023, pp. 7459--7469.

\bibitem{fourier_LLapp3}
Y.~Huang, X.~Tu, G.~Fu, T.~Liu, B.~Liu, M.~Yang, and Z.~Feng, ``Low-light image
  enhancement by learning contrastive representations in spatial and frequency
  domains,'' in \emph{2023 IEEE International Conference on Multimedia and Expo
  (ICME)}.\hskip 1em plus 0.5em minus 0.4em\relax IEEE, 2023, pp. 1307--1312.

\bibitem{fourier_LLapp4}
T.~Zhang, P.~Liu, M.~Zhao, and H.~Lv, ``Dmfourllie: Dual-stage and multi-branch
  fourier network for low-light image enhancement,'' in \emph{Proceedings of
  the 32nd ACM International Conference on Multimedia}, 2024, pp. 7434--7443.

\bibitem{fourier_app3}
H.~Nehete, A.~Monga, P.~Kaushik, and B.~K. Kaushik, ``Fourier prior-based
  two-stage architecture for image restoration,'' in \emph{Proceedings of the
  IEEE/CVF Conference on Computer Vision and Pattern Recognition}, 2024, pp.
  6014--6023.

\bibitem{fourier_app5}
X.~Guo, X.~Fu, M.~Zhou, Z.~Huang, J.~Peng, and Z.-J. Zha, ``Exploring fourier
  prior for single image rain removal.'' in \emph{IJCAI}, 2022, pp. 935--941.

\bibitem{idea1}
C.~Liu, G.~Gao, Z.~Huang, Z.~Hu, Q.~Liu, and Y.~Wang, ``Yolc: You only look
  clusters for tiny object detection in aerial images,'' \emph{IEEE
  Transactions on Intelligent Transportation Systems}, 2024.

\bibitem{idea2}
X.~Yang, J.~Yan, Q.~Ming, W.~Wang, X.~Zhang, and Q.~Tian, ``Rethinking rotated
  object detection with gaussian wasserstein distance loss,'' in
  \emph{International conference on machine learning}.\hskip 1em plus 0.5em
  minus 0.4em\relax PMLR, 2021, pp. 11\,830--11\,841.

\bibitem{LLIE_1}
C.~Wei, W.~Wang, W.~Yang, and J.~Liu, ``Deep retinex decomposition for
  low-light enhancement,'' \emph{arXiv preprint arXiv:1808.04560}, 2018.

\bibitem{LLIE_2}
Y.~Cai, H.~Bian, J.~Lin, H.~Wang, R.~Timofte, and Y.~Zhang, ``Retinexformer:
  One-stage retinex-based transformer for low-light image enhancement,'' in
  \emph{Proceedings of the IEEE/CVF international conference on computer
  vision}, 2023, pp. 12\,504--12\,513.

\bibitem{ret_3}
W.~Wu, J.~Weng, P.~Zhang, X.~Wang, W.~Yang, and J.~Jiang, ``Uretinex-net:
  Retinex-based deep unfolding network for low-light image enhancement,'' in
  \emph{Proceedings of the IEEE/CVF conference on computer vision and pattern
  recognition}, 2022, pp. 5901--5910.

\bibitem{LLIE_3}
X.~Yi, H.~Xu, H.~Zhang, L.~Tang, and J.~Ma, ``Diff-retinex: Rethinking
  low-light image enhancement with a generative diffusion model,'' in
  \emph{Proceedings of the IEEE/CVF International Conference on Computer
  Vision}, 2023, pp. 12\,302--12\,311.

\bibitem{LLIE_4}
A.~Brateanu, R.~Balmez, A.~Avram, C.~Orhei, and C.~Ancuti, ``Lyt-net:
  Lightweight yuv transformer-based network for low-light image enhancement,''
  \emph{IEEE Signal Processing Letters}, 2025.

\bibitem{LLIE_5}
X.~Guo and Q.~Hu, ``Low-light image enhancement via breaking down the
  darkness,'' \emph{International Journal of Computer Vision}, vol. 131, no.~1,
  pp. 48--66, 2023.

\bibitem{LLIE_6}
Q.~Yan, Y.~Feng, C.~Zhang, G.~Pang, K.~Shi, P.~Wu, W.~Dong, J.~Sun, and
  Y.~Zhang, ``Hvi: A new color space for low-light image enhancement,'' in
  \emph{Proceedings of the Computer Vision and Pattern Recognition Conference},
  2025, pp. 5678--5687.

\bibitem{LLIE_7}
T.~Wang, K.~Zhang, T.~Shen, W.~Luo, B.~Stenger, and T.~Lu,
  ``Ultra-high-definition low-light image enhancement: A benchmark and
  transformer-based method,'' in \emph{Proceedings of the AAAI conference on
  artificial intelligence}, vol.~37, no.~3, 2023, pp. 2654--2662.

\bibitem{LLIE_8}
Y.~Jiang, X.~Gong, D.~Liu, Y.~Cheng, C.~Fang, X.~Shen, J.~Yang, P.~Zhou, and
  Z.~Wang, ``Enlightengan: Deep light enhancement without paired supervision,''
  \emph{IEEE transactions on image processing}, vol.~30, pp. 2340--2349, 2021.

\bibitem{fourier_app1}
Z.~Zhou, L.~Qi, and Y.~Shi, ``Generalizable medical image segmentation via
  random amplitude mixup and domain-specific image restoration,'' in
  \emph{European Conference on Computer Vision}.\hskip 1em plus 0.5em minus
  0.4em\relax Springer, 2022, pp. 420--436.

\bibitem{fourier_app2}
Q.~Xu, R.~Zhang, Y.~Zhang, Y.~Wang, and Q.~Tian, ``A fourier-based framework
  for domain generalization,'' in \emph{Proceedings of the IEEE/CVF conference
  on computer vision and pattern recognition}, 2021, pp. 14\,383--14\,392.

\bibitem{fourier_app4}
D.~Fuoli, L.~Van~Gool, and R.~Timofte, ``Fourier space losses for efficient
  perceptual image super-resolution,'' in \emph{Proceedings of the IEEE/CVF
  International Conference on Computer Vision}, 2021, pp. 2360--2369.

\bibitem{fourier_app6}
J.~Huang, Y.~Liu, F.~Zhao, K.~Yan, J.~Zhang, Y.~Huang, M.~Zhou, and Z.~Xiong,
  ``Deep fourier-based exposure correction network with spatial-frequency
  interaction,'' in \emph{European Conference on Computer Vision}.\hskip 1em
  plus 0.5em minus 0.4em\relax Springer, 2022, pp. 163--180.

\bibitem{fourier_LLapp1}
X.~Lv, S.~Zhang, C.~Wang, Y.~Zheng, B.~Zhong, C.~Li, and L.~Nie, ``Fourier
  priors-guided diffusion for zero-shot joint low-light enhancement and
  deblurring,'' in \emph{Proceedings of the IEEE/CVF Conference on Computer
  Vision and Pattern Recognition}, 2024, pp. 25\,378--25\,388.

\bibitem{IG_1}
J.~Ho, A.~Jain, and P.~Abbeel, ``Denoising diffusion probabilistic models,''
  \emph{Advances in neural information processing systems}, vol.~33, pp.
  6840--6851, 2020.

\bibitem{IG_2}
C.~Doersch, ``Tutorial on variational autoencoders,'' \emph{arXiv preprint
  arXiv:1606.05908}, 2016.

\bibitem{se}
J.~Hu, L.~Shen, and G.~Sun, ``Squeeze-and-excitation networks,'' in
  \emph{Proceedings of the IEEE conference on computer vision and pattern
  recognition}, 2018, pp. 7132--7141.

\bibitem{exp5}
X.~Xu, R.~Wang, C.-W. Fu, and J.~Jia, ``Snr-aware low-light image
  enhancement,'' in \emph{Proceedings of the IEEE/CVF conference on computer
  vision and pattern recognition}, 2022, pp. 17\,714--17\,724.

\bibitem{exp8}
Y.~Wang, R.~Wan, W.~Yang, H.~Li, L.-P. Chau, and A.~Kot, ``Low-light image
  enhancement with normalizing flow,'' in \emph{Proceedings of the AAAI
  conference on artificial intelligence}, vol.~36, no.~3, 2022, pp. 2604--2612.

\bibitem{exp6}
Y.~Zhang, J.~Zhang, and X.~Guo, ``Kindling the darkness: A practical low-light
  image enhancer,'' in \emph{Proceedings of the 27th ACM international
  conference on multimedia}, 2019, pp. 1632--1640.

\bibitem{IG_3}
D.~P. Kingma, M.~Welling \emph{et~al.}, ``Auto-encoding variational bayes,''
  2013.

\bibitem{idea3}
S.~Odaibo, ``Tutorial: Deriving the standard variational autoencoder (vae) loss
  function,'' \emph{arXiv preprint arXiv:1907.08956}, 2019.

\bibitem{exp1}
Z.~Fu, Y.~Yang, X.~Tu, Y.~Huang, X.~Ding, and K.-K. Ma, ``Learning a simple
  low-light image enhancer from paired low-light instances,'' in
  \emph{Proceedings of the IEEE/CVF conference on computer vision and pattern
  recognition}, 2023, pp. 22\,252--22\,261.

\bibitem{exp3}
S.~W. Zamir, A.~Arora, S.~Khan, M.~Hayat, F.~S. Khan, and M.-H. Yang,
  ``Restormer: Efficient transformer for high-resolution image restoration,''
  in \emph{Proceedings of the IEEE/CVF conference on computer vision and
  pattern recognition}, 2022, pp. 5728--5739.

\bibitem{exp11}
J.~Bai, Y.~Yin, Q.~He, Y.~Li, and X.~Zhang, ``Retinexmamba: Retinex-based mamba
  for low-light image enhancement,'' in \emph{International Conference on
  Neural Information Processing}.\hskip 1em plus 0.5em minus 0.4em\relax
  Springer, 2024, pp. 427--442.

\bibitem{exp10}
W.~Wang, H.~Yang, J.~Fu, and J.~Liu, ``Zero-reference low-light enhancement via
  physical quadruple priors,'' in \emph{Proceedings of the IEEE/CVF conference
  on computer vision and pattern recognition}, 2024, pp. 26\,057--26\,066.

\bibitem{datasets_LIME}
X.~Guo, Y.~Li, and H.~Ling, ``Lime: Low-light image enhancement via
  illumination map estimation,'' \emph{IEEE Transactions on image processing},
  vol.~26, no.~2, pp. 982--993, 2016.

\bibitem{exp7}
Y.~Zhang, X.~Guo, J.~Ma, W.~Liu, and J.~Zhang, ``Beyond brightening low-light
  images,'' \emph{International Journal of Computer Vision}, vol. 129, pp.
  1013--1037, 2021.

\bibitem{exp2}
S.~W. Zamir, A.~Arora, S.~Khan, M.~Hayat, F.~S. Khan, M.-H. Yang, and L.~Shao,
  ``Learning enriched features for real image restoration and enhancement,'' in
  \emph{Computer Vision--ECCV 2020: 16th European Conference, Glasgow, UK,
  August 23--28, 2020, Proceedings, Part XXV 16}.\hskip 1em plus 0.5em minus
  0.4em\relax Springer, 2020, pp. 492--511.

\bibitem{exp4}
W.~Yang, W.~Wang, H.~Huang, S.~Wang, and J.~Liu, ``Sparse gradient regularized
  deep retinex network for robust low-light image enhancement,'' \emph{IEEE
  Transactions on Image Processing}, vol.~30, pp. 2072--2086, 2021.

\bibitem{exp9}
Y.~Liang, B.~Wang, W.~Ren, J.~Liu, W.~Wang, and W.~Zuo, ``Learning hierarchical
  dynamics with spatial adjacency for image enhancement,'' in \emph{Proceedings
  of the 30th ACM International Conference on Multimedia}, 2022, pp.
  2767--2776.

\bibitem{datasets_LSRWHuawei}
J.~Hai, Z.~Xuan, R.~Yang, Y.~Hao, F.~Zou, F.~Lin, and S.~Han, ``R2rnet:
  Low-light image enhancement via real-low to real-normal network,''
  \emph{Journal of Visual Communication and Image Representation}, vol.~90, p.
  103712, 2023.

\bibitem{datasets_DICM}
C.~Lee, C.~Lee, and C.-S. Kim, ``Contrast enhancement based on layered
  difference representation of 2d histograms,'' \emph{IEEE transactions on
  image processing}, vol.~22, no.~12, pp. 5372--5384, 2013.

\bibitem{datasets_MEF}
K.~Ma, K.~Zeng, and Z.~Wang, ``Perceptual quality assessment for multi-exposure
  image fusion,'' \emph{IEEE Transactions on Image Processing}, vol.~24,
  no.~11, pp. 3345--3356, 2015.

\bibitem{datasets_NPE}
S.~Wang, J.~Zheng, H.-M. Hu, and B.~Li, ``Naturalness preserved enhancement
  algorithm for non-uniform illumination images,'' \emph{IEEE transactions on
  image processing}, vol.~22, no.~9, pp. 3538--3548, 2013.

\bibitem{datasets_VV}
V.~Vonikakis, R.~Kouskouridas, and A.~Gasteratos, ``On the evaluation of
  illumination compensation algorithms,'' \emph{Multimedia Tools and
  Applications}, vol.~77, pp. 9211--9231, 2018.

\end{thebibliography}

\end{document}